\documentclass[10pt,twocolumn,letterpaper]{article}

\usepackage[pagenumbers]{wacv}

\usepackage[normalem]{ulem}
\newcommand{\keeptext}[1]{#1}
\newcommand{\droptext}[1]{}

\usepackage[accsupp]{axessibility}

\definecolor{wacvblue}{rgb}{0.21,0.49,0.74}
\usepackage[pagebackref,breaklinks,colorlinks,allcolors=wacvblue]{hyperref}

\def\confName{WACV}
\def\confYear{2027}

\usepackage{bbm}
\usepackage{algorithm}
\usepackage{algorithmic}

\title{SymAnt: Grounding Long-Horizon Action Anticipation with Adaptable Symbolic Priors}

\author{
Fnu Aryan$^{1}$ \quad
Yaqi Xie$^{1}$ \quad
Joseph Campbell$^{3}$ \quad
Woojun Kim$^{1}$ \quad
Kwonjoon Lee$^{2}$\\
Hossein Nourkhiz Mahjoub$^{2}$ \quad
Ehsan Moradi-Pari$^{2}$ \quad
Katia Sycara$^{1}$ \quad
Simon Stepputtis$^{4}$\\[4pt]
$^{1}$Carnegie Mellon University \quad
$^{2}$Honda Research Institute USA, Inc.\\
$^{3}$Purdue University \quad
$^{4}$Virginia Tech\\[2pt]
{\tt\small aryanmangal2022@gmail.com, yaqixie@cmu.edu, joecamp@purdue.edu, woojunk@andrew.cmu.edu}\\
{\tt\small kwonjoon\_lee@honda-ri.com, hossein\_nourkhizmahjoub@honda-ri.com, emoradipari@honda-ri.com}\\
{\tt\small sycara@andrew.cmu.edu, stepputtis@vt.edu}
}

\begin{document}
\maketitle
\begin{abstract}
As robots and intelligent systems increasingly interact with humans, they must anticipate long sequences of future actions from only brief observations.
While purely neural models typically require extensive observations to predict future actions and remain difficult to interpret and adjust, symbolic priors offer a structured alternative.
To this end, we introduce SymAnt, a novel neuro-symbolic approach that unifies the expressiveness of neural networks with 
the structured reasoning of scene and knowledge graphs.
Given a short video, SymAnt performs a joint-search over both graphs to construct a symbolic context, enabling unified reasoning over spatial relationships, object attributes, affordances, and associated actions. 
This symbolic context serves as a prior for an iterative refinement process that progressively generates and improves the sequence of future actions.
By grounding predictions in symbolic context, SymAnt anticipates long-horizon actions from short observation windows in a parameter-efficient manner, while enabling users to directly edit the underlying knowledge to adjust predictions without further training. 
Through our experiments, we demonstrate the effectiveness of SymAnt, particularly when only short observation contexts are available, across \droptext{}\keeptext{five} diverse datasets: \textit{Breakfast}, \textit{50 Salads}, \textit{EPIC-KITCHENS}, \droptext{} \textit{EGTEA Gaze+}\keeptext{, and \textit{Assembly101}}.
\end{abstract}

\section{Introduction}
\label{sec:intro}

\begin{figure*}[htbp]
    \centering
    \includegraphics[width=1\linewidth]{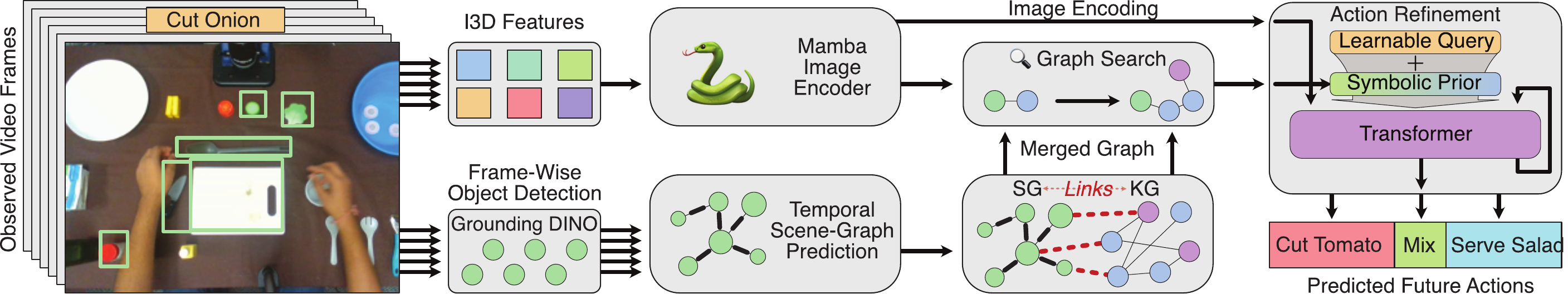}
    \caption{SymAnt architecture: Utilizing a short video of prior observations, we extract a relevant scene graph and utilize a knowledge graph as additional context to predict a sequence of future actions. We propose a joint-graph search to capture additional context, utilizing it as a prior for an iterative action refinement.}
    \label{fig:overview}
\end{figure*}

Action anticipation is crucial for intelligent agents in tasks such as human-robot collaboration~\cite{Buerkle2021, Gammulle2019}, autonomous driving~\cite{Guo2024, Leung2020}, and reconnaissance~\cite{Banijamali2021, Lai2024}.
Accurately anticipating the actions and behavior of other agents, including humans, allows systems to adjust their own behavior proactively. 
Particularly in real-world applications, being able to predict actions from \textit{short} observation contexts is critical.
At the same time, correctly predicting future actions over \textit{long} horizons is essential for various downstream tasks involving human-agent interaction, such as collaborative cooking (see~\cref{fig:overview}).
However, current neural-only action anticipation methods~\cite{gong2022future, zhong2023diffant} require large amounts of training data, long observation contexts for inference, and do not easily adapt to changing task dynamics. 
To address this challenge, we propose the integration of symbolic knowledge as an additional context for action anticipation, leveraging both scene and knowledge graphs for improved contextual understanding.  

Specifically, we introduce a method that represents this context as a scene graph (SG) to capture spatial relationships, as well as a knowledge graph (KG) that augments the SG with attributes, affordances, and related actions of the objects detected in the environment. 
A key contribution of our work is the joint-graph search, which dynamically merges both graphs to enable explicit reasoning: 
For example, a \textit{tomato} node in the KG is connected to an affordance node \textit{cuttable}, while the SG represents a \textit{knife}, with the affordance \textit{can cut}, located \textit{next to} it. 
In this example, a likely future action is \textit{cut tomato}. 
This combined graph search provides an additional information-rich context which compensates for the lack of visual history in short observation windows, enabling accurate long-horizon anticipation. 
As predictions are grounded in this explicit representation, our model remains parameter efficient while allowing users to edit the knowledge to easily change the model's predictions. 

Given this structured context, we frame action anticipation as a deterministic iterative refinement task, where the predicted sequence of future actions is progressively generated and refined starting from the symbolic knowledge retrieved via our joint-graph search. Architecturally, we leverage learned continuous representations of observed video alongside our retrieved prior as initialization for the refinement process, ultimately outputting discrete action sequences.  
This iterative mechanism allows the model to globally review and adjust the sequence, ensuring logical consistency at longer horizons. In contrast to diffusion-inspired approaches~\cite{zhong2023diffant, gong2024actfusion, 10081412} that rely on stochastic noise, our symbolic initialization approach utilizes a deterministic, structured, and semantically grounded starting point, enabling accurate anticipation even when visual history is sparse. 
In summary, we contribute the following:
\begin{itemize}
    \item We introduce SymAnt, a neuro-symbolic framework that fuses scene and knowledge graphs via joint-graph search, providing structured symbolic context for long-horizon action anticipation from short observation windows.
    \item We show that leveraging a symbolic context enables runtime adaptability, directly influencing the predicted actions based on graph changes.
    \item We evaluate SymAnt across \droptext{}\keeptext{five} benchmarks (\textit{Breakfast}, \textit{50 Salads}, \textit{EPIC-KITCHENS}, \textit{EGTEA Gaze+}\keeptext{, \textit{Assembly101}}), achieving state-of-the-art performance on short-observation settings while maintaining competitive results on longer contexts and retaining a small model. 

\end{itemize}
\section{Related Work}
\subsection{Symbolic Knowledge for Computer Vision}
The integration of structured domain knowledge, particularly through knowledge graphs~\cite{ji2021survey}, into vision models is significantly enhancing grounding, interpretability, and performance~\cite{marino2016more, zhu2014capturing, malisiewicz2009beyond, li2023airloc}. 
Neuro-symbolic vision pipelines~\cite{jiao2022graph} leverage the structure and hierarchy of knowledge graphs to enhance vision tasks, including object detection~\cite{fang2017object, lang2022robust}, scene graph generation~\cite{zareian2020bridging}, transfer learning~\cite{monka2022survey, chen2020knowledge}, vision-language pre-training~\cite{alberts2020visualsem}, classification~\cite{zhang2019knowledge, tang2023selective}, and image captioning~\cite{zhou2019improving, zhao2023boosting}.
For example,~\cite{marino2016more} introduces a graph search mechanism over knowledge graphs, and subsequent work~\cite{bhagat2023sample} extended this to include novel concepts. \droptext{}
\keeptext{Recent advances further optimize this graph reasoning; DiffLogic~\cite{chen2023difflogic} introduces differentiable rule-based inference over large knowledge graphs, while Ontiveros et al.~\cite{ontiveros2025grounding} highlight the critical role of the grounding criterion during instantiation.}
In parallel, scene graphs have also been extensively utilized in computer vision for tasks such as object detection, image captioning, and visual question answering~\cite{chang2021comprehensive, li2017scene, hildebrandt2020scene, cong2018scene, mozes2021scene}. In the realm of action anticipation, scene graphs have been employed to predict future actions by modeling object interactions~\cite{rodin2024action}. 
\keeptext{Additionally, neuro-symbolic models like LASER~\cite{huang2025laser} and NAVER~\cite{cai2025naver} employ differentiable reasoning and explicit logic for scene-graph generation and visual grounding.}
While previous methods used knowledge graphs or scene graphs independently, we integrate both, reasoning over object knowledge, instance counts, and spatial relationships.
To our knowledge, our approach is the first to leverage \droptext{} \keeptext{an explicit joint search (with a learned importance network) over} scene and knowledge graphs \droptext{} for action anticipation. 

\subsection{Action Anticipation}
\droptext{}\keeptext{Action anticipation is the task of predicting future actions from observed video of humans or agents performing a task.}
\droptext{}\keeptext{Early work focused on third-person video~\cite{abu2018will, gao2017red, gong2022future, ke2019time}; recent benchmarks~\cite{damen2018scaling, damen2020epic,grauman2022ego4d} have shifted attention to first-person (egocentric) vision.} 
Early methods for long-term action anticipation used RNNs~\cite{abu2018will, furnari2019would} and TCNs~\cite{abu2021long} to model action progressions over time. More recently, Transformer-based models~\cite{girdhar2021anticipative, gong2022future, nawhal2022rethinking, zhong2023anticipative} have demonstrated superior performance in learning long-range temporal relationships, making them the preferred choice for long-term action anticipation.
Recent work~\cite{zhong2023diffant, gong2024actfusion, wang2023pdpp} has used diffusion models to generate latent embeddings of discrete actions. 
\droptext{}
\keeptext{Recent approaches increasingly incorporate structural and semantic priors. One line of work injects this structure implicitly: modeling hand--object interactions~\cite{roy2024interaction}, inferring latent goal distributions~\cite{roy2024abstractgoal}, or distilling language-model semantics into visual prototypes~\cite{diko2024semantically}. In contrast, neuro-symbolic approaches introduce explicit structure. For example, NeSCA~\cite{bhagat2023knowledgeguided} leverages static knowledge graphs for short-context anticipation. Unlike both implicit methods and prior neuro-symbolic works, we introduce a dynamic joint-graph between scene and knowledge graphs. This explicit prior is iteratively refined and remains fully editable during deployment, allowing us to reduce parameters and the need for extensive observation contexts.}
\section{Methodology}

In this section, we introduce our method, \textit{SymAnt} (\cref{fig:overview}), a novel approach for action anticipation that utilizes symbolic domain knowledge as additional context to improve prediction through an iterative action decoder. 
The central goal of our approach is to bootstrap information from the scene and knowledge graph to inform the generation of plausible action sequences. 
This symbolic grounding reduces the required observation context while allowing the model's behavior to be adjusted post-training through edits to the contextual knowledge.

\subsection{Problem Statement}\label{sec:problem}
Given a set of features from an observed video context ($V_o$), $\mathbf{F} \in \mathbb{R}^{L \times G}$ with $G$ dimensions across $L$ frames, the objective of the anticipation task is to predict the subsequent future actions over a specific time horizon following the observed video.
These actions consist of action classes $\mathbf{A} \in \mathbb{R}^{N \times C}$ and their corresponding durations $\mathbf{d} \in \mathbb{R}^{N \times 1}$, where $C$ represents the total number of action classes.
On a high level, SymAnt utilizes a Mamba-based encoder $(\mathbf{E} = f_{enc}(\mathbf{F}))$ to encode the video observations and a novel iterative action decoder $f_{dec}(\dots)$, initialized with learned prior $f_{SN}(\dots)$ from the temporal scene graph and symbolic knowledge. Pseudocode for all algorithms is provided in supplementary material.

\subsection{Symbolic Context Aggregation}
\label{sec:gsnn}

The key to our approach is utilizing symbolic knowledge to derive additional contextual information from the objects that can be seen in a short-context video observation. 
We achieve this by creating a \textit{joint-graph} from a scene graph based on the video observations $V_o$ and a fixed knowledge graph built once from the training split, $\mathbf{K} = \{\mathbf{K}_n, \mathbf{K}_A\}$ providing additional information for the detected entities, where $\mathbf{K}_n, \mathbf{K}_A$ are nodes and adjacency matrix of $\mathbf{K}$, respectively. We construct our knowledge graph to capture the affordances and action dependencies relevant to our experiments (see experiments on KG source in~\cref{sec:kg_ablation}). %
Our knowledge graph is built \keeptext{automatically} by (i) extracting objects from the dataset annotations, (ii) linking objects to their functional affordances (e.g., \textit{bowl} $\rightarrow$ \textit{hold}), and (iii) linking commonsense object-object relations such as containment, proximity, and temporal co-occurrence derived directly from the dataset
\droptext{}
\keeptext{\textbf{training split only}, restricted to the observed portion of each video, so neither test data nor future-action labels contribute at any stage. The KG is built once before training and is never updated at inference.}
Our context aggregation module operates in three key steps given the observed video sequence and a knowledge graph, which are explained as follows: 

1. \textbf{Temporal Scene Graph Generation}: Scene graphs excel at capturing spatial relationships between entities; however, for action anticipation, we modify them to capture temporal changes as well.
We start by utilizing Grounding DINO~\droptext{}\keeptext{\cite{liu2024grounding}} on each of observed frames $V_o^l$ to detect initial concepts \( \mathbf{c}^l \) in the scene, where $l$ is the frame number.
This forms the basis of our frame-level scene graphs \( \mathbf{S}^l \). 
The spatial relationship between two scene entities is predicted with a neural network \(f_{\text{edge}}() \), classifying whether or not two nodes have a spatial relationship, and if so, estimates the relationship type with a respective weight, indicating its importance.
The set of spatial relations are: `\textit{on}', `\textit{next to}', `\textit{behind}', `\textit{in front of}', `\textit{above}', `\textit{across}', `\textit{below}', `\textit{inside}', `\textit{under}', `\textit{left}', `\textit{right}', `\textit{in}'.
\keeptext{$f_{\text{edge}}$ is trained on Visual Genome~\cite{krishna2017visual}, using the scenes annotated with these relations, to avoid any form of leakage.}
Concepts in the SG are represented as $f_{ViT}(box)$~\cite{dosovitskiy2020image}, a latent representation of entity within the bounding box. To aggregate scene graphs \( \mathbf{S}^l \) across multiple time frames and construct a temporal scene graph \( \mathbf{S} \), we adopt a dynamic update mechanism. For each new frame $l$, any object not already present in the temporal scene graph is added, along with its relationships to the existing nodes. When a relationship between two pre-existing nodes is updated, a new relationship is added to the graph while retaining but decaying the importance of previous one, to prevent accumulation of stale information.

2. \textbf{Graph Merging Module}: The goal of this module is to combine scene-specific information from the temporal scene graph \( \mathbf{S} \) with domain knowledge from the provided knowledge graph \( \mathbf{K} \) based on the common entities present in the environment. 
Given a temporal SG $\mathbf{S} = \{\mathbf{S}_n, \mathbf{S}_A\}$, where $\mathbf{S}_n$ and $\mathbf{S}_A$ are the nodes and adjacency matrix of $\mathbf{S}$ respectively, we form connections between the graphs by linking the mutual nodes between the scene and knowledge graph to create the joint-graph  $\mathbf{M} = \{\mathbf{M} _n, \mathbf{M} _A\}$, which includes the entire scene and knowledge graph. In scenarios where the scene graph contains multiple instances of the same node, all such instances are connected to the corresponding single node in the knowledge graph. 
The nodes that link the scene and knowledge graphs are referred to as active nodes $\mathbf{M}_{\text{active}}$. These nodes serve as the starting points for our joint-graph search
(see the next section for details about graph search). 

\begin{table*}[htbp]
\centering
\caption{Comparison of anticipation performance with short observations ($\alpha = 0.05, 0.1$) on Breakfast and 50 Salads datasets. SymAnt is best in every setting, by $3.2\%$ on average over NeSCA. \keeptext{Cells marked $-$ for the LLM baselines were not evaluated under our inference budget.} \keeptext{The last row reports the standard deviation of our results over the official splits.}}
\label{short_context_bf_50s}

\resizebox{\textwidth}{!}{
\begin{tabular}{l cccc cccc cccc cccc}
\toprule
& \multicolumn{4}{c}{\textbf{Breakfast} ($\alpha=0.05$)} & \multicolumn{4}{c}{\textbf{Breakfast} ($\alpha=0.1$)} & \multicolumn{4}{c}{\textbf{50 Salads} ($\alpha=0.05$)} & \multicolumn{4}{c}{\textbf{50 Salads} ($\alpha=0.1$)} \\
\cmidrule(lr){2-5} \cmidrule(lr){6-9} \cmidrule(lr){10-13} \cmidrule(lr){14-17}
Method & $\beta=0.1$ & 0.2 & 0.3 & 0.5 & 0.1 & 0.2 & 0.3 & 0.5 & 0.1 & 0.2 & 0.3 & 0.5 & 0.1 & 0.2 & 0.3 & 0.5 \\
\midrule
KG \cite{bhagat2023sample}  & 5.44 & 4.95 & 4.22 & 3.98 & 6.02 & 5.15 & 4.86 & 4.51 & 6.92 & 6.21 & 6.01 & 5.58 & 7.13 & 6.48 & 6.07 & 5.78 \\
\droptext{}
RNN \cite{abu2018will} & 6.16 & 5.60 & 5.53 & 4.96 & 7.67 & 6.73 & 6.15 & 5.22 & 7.98 & 6.90 & 6.48 & 6.42 & 8.78 & 7.92 & 7.57 & 7.26 \\
DS-r1 \cite{guo2025deepseek} & 5.34 & 8.25 & 5.00 & 6.23 & 7.41 & 10.20 & 7.06 & 8.09 & 6.67 & 5.36 & 4.04 & 2.80 & 10.20 & 6.92 & 4.95 & 3.30 \\
FUTR \cite{gong2022future} & 9.54 & 7.24 & 6.42 & 5.58 & 14.70 & 12.55 & 12.10 & 11.71 & 8.90 & 7.46 & 7.29 & 8.63 & 15.17 & 11.34 & 11.31 & 11.36 \\
GPT-o1 \cite{openai2024o1} & 7.61 & 6.67 & 7.89 & 5.15 & $-$ & $-$ & $-$ & $-$ & 15.64 & 13.12 & 13.38 & 11.18 & 15.18 & 15.07 & 13.47 & 12.29 \\
Qwen3-Omni \cite{qwen2025omni} & $-$ & $-$ & $-$ & $-$ & $-$ & $-$ & $-$ & $-$ & $-$ & $-$ & $-$ & $-$ & 21.45 & \underline{18.78} & 16.53 & 12.97 \\
NeSCA \cite{bhagat2023knowledgeguided} & \underline{9.91} & \underline{7.95} & \underline{6.86} & \underline{5.88} & \underline{15.53} & \underline{13.52} & \underline{13.07} & \underline{11.94} & \underline{17.86} & \underline{16.25} & \underline{10.84} & \underline{9.38} & \underline{23.15} & 17.28 & \underline{16.62} & \underline{13.96} \\
Ours & \textbf{12.19} & \textbf{9.80} & \textbf{8.91} & \textbf{8.15} & \textbf{17.11} & \textbf{14.32} & \textbf{14.06} & \textbf{12.92} & \textbf{21.98} & \textbf{19.41} & \textbf{17.75} & \textbf{14.33} & \textbf{27.92} & \textbf{23.82} & \textbf{20.01} & \textbf{18.83} \\
\keeptext{Ours ($\pm$std)} & \keeptext{1.7} & \keeptext{0.4} & \keeptext{1.1} & \keeptext{0.4} & \keeptext{0.8} & \keeptext{0.5} & \keeptext{1.5} & \keeptext{0.7} & \keeptext{2.2} & \keeptext{2.7} & \keeptext{1.5} & \keeptext{3.1} & \keeptext{1.6} & \keeptext{3.5} & \keeptext{2.2} & \keeptext{3.2} \\
\bottomrule
\end{tabular}
} %
\end{table*}
3. \textbf{Joint-Graph Search}: The goal of the joint-graph search is to iteratively expand around the currently active nodes, utilizing both spatial and semantic information to gather comprehensive context.
The expansion continues until all relevant domain information is obtained, i.e., when no additional nodes contribute meaningful context to the current observation. Our joint-graph search uses a three-stage approach inspired by~\cite{bhagat2023sample, aryan2024symbolic, marino2016more}, which comprises:
\begin{itemize}
    \item \textbf{Propagation Network}: The propagation network, $f_{prop}(\dots)$ updates the states of active nodes \( \mathbf{M}_{active} \), as well as the states of neighboring nodes that are connected to these active nodes given the context of the video $\mathbf{F}$.
    \item \textbf{Importance Network}: The Importance Network, $f_{imp}(\dots)$ 
    decides whether adjacent nodes to currently active ones should be activated, thereby preventing exponential growth of the graph. The number of iterations of alternating between propagation and importance network is dynamically determined (see supp. Alg.~4)
    \item \textbf{Context Network}: The Context network, $f_{cont}(\dots)$ generates a final embedding $\mathbf{z}_T$ using active nodes in the expanded graph, used as context for the decoder.  
\end{itemize}

\subsection{Video Encoder}

In addition to the symbolic knowledge derived from the observed video,
we extract features \droptext{}from the video frames using I3D~\cite{carreira2017quo}, which employs 3D convolutions to learn rich motion and appearance representations. 
These input features $\mathbf{F}$ are \droptext{}transformed using the Mamba~\cite{gu2023mamba} model as it improves training and inference speed without compromising accuracy compared to transformer encoders \droptext{}in FUTR~\cite{gong2022future} and DiffAnt~\cite{zhong2023diffant} (see ablations for details).
We combine image features with a sinusoidal positional encoding $\mathbf{P} \in \mathbb{R}^{L \times D}$, enabling the Mamba encoder to generate refined representations $\mathbf{E} \in \mathbb{R}^{L \times D}$. 
These representations are then processed through a linear classifier to map them to the discrete actions, enabling recognition of observed actions in videos, denoted as \(\tilde{\mathbf{A}}^{\text{obs}}\). Training these representations for action recognition ensures that they capture meaningful temporal and spatial structures, also providing robustness against noisy object detections.
This representation is used as conditioning for the iterative action refinement process, as explained in the next section.

\begin{table*}[htbp]
\centering
\caption{Comparison of long-term anticipation performance with observation horizons $\alpha=0.2, 0.3$ on the Breakfast and 50 Salads datasets. Results show that SymAnt excels at longer prediction horizons ($\beta=0.5$), with largest margins on 50 Salads.}
\label{long_context_bf_50s}

\resizebox{\textwidth}{!}{
\begin{tabular}{l cccc cccc cccc cccc}
\toprule
& \multicolumn{4}{c}{\textbf{Breakfast} ($\alpha=0.2$)} & \multicolumn{4}{c}{\textbf{Breakfast} ($\alpha=0.3$)} & \multicolumn{4}{c}{\textbf{50 Salads} ($\alpha=0.2$)} & \multicolumn{4}{c}{\textbf{50 Salads} ($\alpha=0.3$)} \\
\cmidrule(lr){2-5} \cmidrule(lr){6-9} \cmidrule(lr){10-13} \cmidrule(lr){14-17}
Method & $\beta=0.1$ & 0.2 & 0.3 & 0.5 & 0.1 & 0.2 & 0.3 & 0.5 & 0.1 & 0.2 & 0.3 & 0.5 & 0.1 & 0.2 & 0.3 & 0.5 \\
\midrule
DS-r1 \cite{guo2025deepseek} & 15.77 & 11.94 & 10.61 & 6.60 & 16.40 & 13.13 & 9.87 & 6.08 & 7.83 & 5.87 & 3.76 & 2.30 & 6.00 & 4.99 & 3.56 & 2.39 \\
\droptext{}
RNN \cite{abu2018will} & 18.11 & 17.20 & 15.94 & 15.81 & 21.64 & 20.02 & 19.73 & 19.21 & 30.06 & 25.43 & 18.74 & 13.49 & 30.77 & 17.19 & 14.79 & 09.77\\
GPT-o1 \cite{openai2024o1} & $-$ & $-$ & $-$ & $-$ & $-$ & $-$ & $-$ & $-$ & 21.60 & 21.75 & 17.86 & 12.93 & 28.38 & 25.08 & 20.44 & 15.99\\
UAAA \cite{abu2019uncertainty} & 16.71 & 15.40 & 14.47 & 14.20 & 20.73 & 18.27 & 18.42 & 16.86 & 24.86 & 22.37 & 19.88 & 12.82 & 29.10 & 20.50 & 15.28 & 12.31 \\
\droptext{}
TimeCond. \cite{ke2019time} & 18.41 & 17.21& 16.42& 15.84& 22.75& 20.44& 19.64& 19.75& 32.51 & 27.61 & 21.26 & 15.99 & 35.12 & 27.05 & 22.05 & 15.59 \\
CycleCons \cite{abu2021long} & 25.88 & 23.42 & 22.42 & 21.54 & 29.66 & 27.33 & 25.58 & 25.20 & 34.76 & 28.41 & 21.82 & 15.25 & 34.39 & 23.70 & 18.95 & 15.89 \\
\droptext{}
FUTR \cite{gong2022future} & 27.70 & 24.55 & 22.83 & 22.04 & 32.27 & 29.88 & 27.49 & 25.87 & 39.55 & 27.54 & 23.31 & 17.77 & 35.15 & 24.86 & 24.22 & 15.26\\
ObjPrompt \cite{zhang2024object} & $-$ & $-$ & $-$ & $-$ & $-$ & $-$ & $-$ & $-$ & 37.40 & 28.90 & 24.20 & 18.10 & 28.00 & 24.00 & 24.30 & 19.30\\
ActFusion \cite{gong2024actfusion} & 28.25 & 25.52 & 24.66 & 23.25 & 35.79 & 31.76 & 29.64 & 28.78 & 39.55 & 28.60 & 23.61 & 19.90 & \underline{42.80} & 27.11 & 23.48 & \underline{22.07}\\
\keeptext{S-GEAR \cite{diko2024semantically}} & \keeptext{$-$} & \keeptext{$-$} & \keeptext{$-$} & \keeptext{$-$} & \keeptext{$-$} & \keeptext{$-$} & \keeptext{$-$} & \keeptext{$-$} & \keeptext{41.0} & \keeptext{28.5} & \keeptext{21.5} & \keeptext{15.3} & \keeptext{41.0} & \keeptext{27.8} & \keeptext{21.4} & \keeptext{16.7} \\
Diffant \cite{zhong2023diffant} & 25.33 & 24.59 & 24.39 & 22.74 & 32.13 & 31.83 & 31.18 & \textbf{30.77} & 36.13 & \underline{34.00} & \underline{30.46} & \underline{25.29} & 34.09 & \underline{30.14} & \underline{26.34} & 20.23\\
Con-Alig. \cite{patsch2024long} & 28.00 & 24.90 & 23.10 & 22.10 & 33.40 & 30.80 & 29.00 & 26.50 & 41.10 & 31.40 & 24.80 & 19.20 & 35.80 & 27.50 & 23.30 & 18.10 \\
LLMAction \cite{wang2024llmaction} & 28.60 & 25.23 & 25.27 & 24.26 & 34.30 & 31.35 & 28.71 & 27.81 & 32.30 & 28.75 & 22.94 & 18.52 & 37.97 & 24.34 & 24.27 & 15.02 \\
ActionLLM \cite{wang2025multimodal} & 26.38 & 23.32 & 21.70 & 20.43 & 31.77 & 29.45 & 26.68 & 24.55 & \textbf{43.67} & 32.80 & 25.73 & 19.51 & \textbf{44.95} & 29.06 & 23.15 & 18.37 \\
MANTA \cite{zatsarynna2025manta} & 27.70 & 25.30 & 24.60 & 23.80 & 34.20 & 30.90 & 29.10 & 27.70 & 28.60 & 22.80 & 19.50 & 13.60 & 31.30 & 21.90 & 17.60 & 13.00 \\
TCCA \cite{mate2024temporal} & \underline{28.83} & 25.93 & 25.19 & 23.41 & \textbf{36.69} & \textbf{33.39} & \textbf{31.96} & 27.63 & \underline{41.99} & 33.69 & 25.85 & 21.96 & 37.12 & 28.31 & 22.85 & 18.43 \\
MixANT \cite{wasim2025mixant} & \textbf{29.6} & \underline{26.3} & \underline{25.9} & \underline{25.0} & \underline{36.2} & \underline{32.8} & \underline{31.2} & 28.7 & 30.3 & 25.0 & 20.9 & 15.2 & 33.4 & 23.7 & 19.7 & 14.6 \\
Ours & 28.22 & \textbf{27.55} & \textbf{27.02} & \textbf{25.67} & 34.22 & 32.06 & 31.19 & \underline{30.11} & 40.78 & \textbf{35.08} & \textbf{31.19} & \textbf{28.19} & 35.23 & \textbf{30.94} & \textbf{29.34} & \textbf{25.50} \\
\keeptext{Ours ($\pm$std)} & \keeptext{1.8} & \keeptext{0.8} & \keeptext{1.1} & \keeptext{0.6} & \keeptext{0.7} & \keeptext{1.6} & \keeptext{2.0} & \keeptext{1.5} & \keeptext{5.3} & \keeptext{1.8} & \keeptext{1.5} & \keeptext{5.1} & \keeptext{4.2} & \keeptext{4.6} & \keeptext{2.1} & \keeptext{2.0} \\
\bottomrule
\end{tabular}
} %
\end{table*}
\subsection{Iterative Symbolic Action Generation}
\label{sec:decoder}
Our action decoder combines the complementary strengths of high-level abstract information encoded in the fully expanded joint-graph and the fine-grained visual details present in the video stream. 
Our approach is built on the hypothesis that the context embedding \(\mathbf{z}_T\) derived from the fully expanded graph inherently contains a noisy representation of future events. 
Consequently, the decoder, utilizing a transformer backbone, iteratively refines this noisy representation into coherent predictions in $T$ steps.
Importantly, each refinement is conditioned on the video features, ensuring that the predictions are semantically meaningful and closely aligned with the observed visual content.

To facilitate focused reasoning for each action while maintaining their interdependencies, we employ a query-based approach~\cite{carion2020end,gong2022future}.
The action queries, denoted as $\mathbf{Q} \in \mathbb{R}^{N \times D}$, consist of $N$ learnable tokens temporally aligned with future action sequences, with each query corresponding to a specific action \droptext{}.
These queries are not fixed representations; rather, they are \keeptext{randomly initialized and} learned dynamically \keeptext{by backpropagation, as in FUTR~\cite{gong2022future}}, allowing the model to refine their embeddings throughout training. These action queries are \droptext{} \keeptext{added element-wise to the} context embedding $\mathbf{z}_T$ from joint-graph-search to form the input for action decoder at initial refinement step $T$ \keeptext{(see Alg.~4 in the supplementary material)}.

To ensure effective learning of the refinement process, we apply supervision at each refinement step. At each refinement step \( t \), the model transforms the action embeddings \( \mathbf{z}_t \) into intermediate action outputs \( \tilde{\mathbf{A}}_t \in \mathbb{R}^{N \times C} \) and timestamp outputs \( \tilde{\mathbf{d}}_t \in \mathbb{R}^{N \times 1} \) using an output projection module. This module comprises two fully connected multi-layer perceptrons\droptext{}\keeptext{: the first is followed by a softmax over the $C$ action classes, while the second regresses the duration of each query directly}. By aligning these intermediate outputs with the ground truth, the model progressively refines its predictions rather than depending solely on the final output. After $T$ refinement steps, the final prediction ${\mathbf{z}_{0}}$ is processed by the same output projection module to generate the final action labels $\tilde{\mathbf{A}} $ and durations $\tilde{\mathbf{d}}$, which serve as the model’s outputs. The exact mathematical formulations for our overall training objective and individual loss components are provided in the supplementary material.

\section{Experiments}
In this section, we evaluate our approach on 
\droptext{}\keeptext{five} widely used action anticipation datasets: two third-person datasets, 50 Salads~\cite{stein2013combining} and Breakfast~\cite{kuehne2014language}, and two egocentric datasets, EPIC-KITCHENS~\cite{damen2020epic} and EGTEA Gaze+~\cite{li2018eye}\keeptext{, as well as the non-cooking Assembly101~\cite{sener2022assembly101}}, and compare the results with neuro-symbolic and neural-only baselines, including fine-tuned and non fine-tuned large language model baselines.
\keeptext{For non-finetuned VLMs, we prompt with observed frames, whereas for LLMs we give the ground-truth action labels as input (see supplementary for the full protocol).}
\keeptext{In these comparisons, SymAnt uses no action labels as input: objects are obtained with an off-the-shelf detector, and the symbolic prior comes from a fixed knowledge graph.}
Additionally, we present qualitative results on a custom real-world kitchen setup and conduct ablation studies to analyze the contributions of different components of our approach. 

\subsection{Short Context Action Anticipation}
We present our experimental results focusing on short observation contexts ($\alpha = 0.05, 0.1$) for long-horizon action anticipation ($\beta = 0.1, 0.2, 0.3, 0.5$) on the 50 Salads and Breakfast datasets in Table~\ref{short_context_bf_50s}, where $\alpha$ is the ratio of observed to total video length and $\beta$ is ratio of predicted to total video length.
We compare SymAnt with neuro-symbolic methods (e.g., NeSCA~\cite{bhagat2023knowledgeguided}), neural-only approaches~\cite{gong2022future, abu2018will}, a purely symbolic KG-Baseline~\cite{bhagat2023sample}, and reasoning models, including ChatGPT o1~\cite{openai2024o1}, Deepseek-r1~\cite{guo2025deepseek}, and Qwen3-Omni~\cite{qwen2025omni}. 
In the short-context anticipation task, bootstrapping symbolic knowledge significantly improves future action prediction, surpassing the closest baseline by $3.2\%$ and LLM ones by $5$–$10\%$ MoC on average. 
\begin{figure*}[htbp]
\centering
\includegraphics[width=0.9\linewidth]{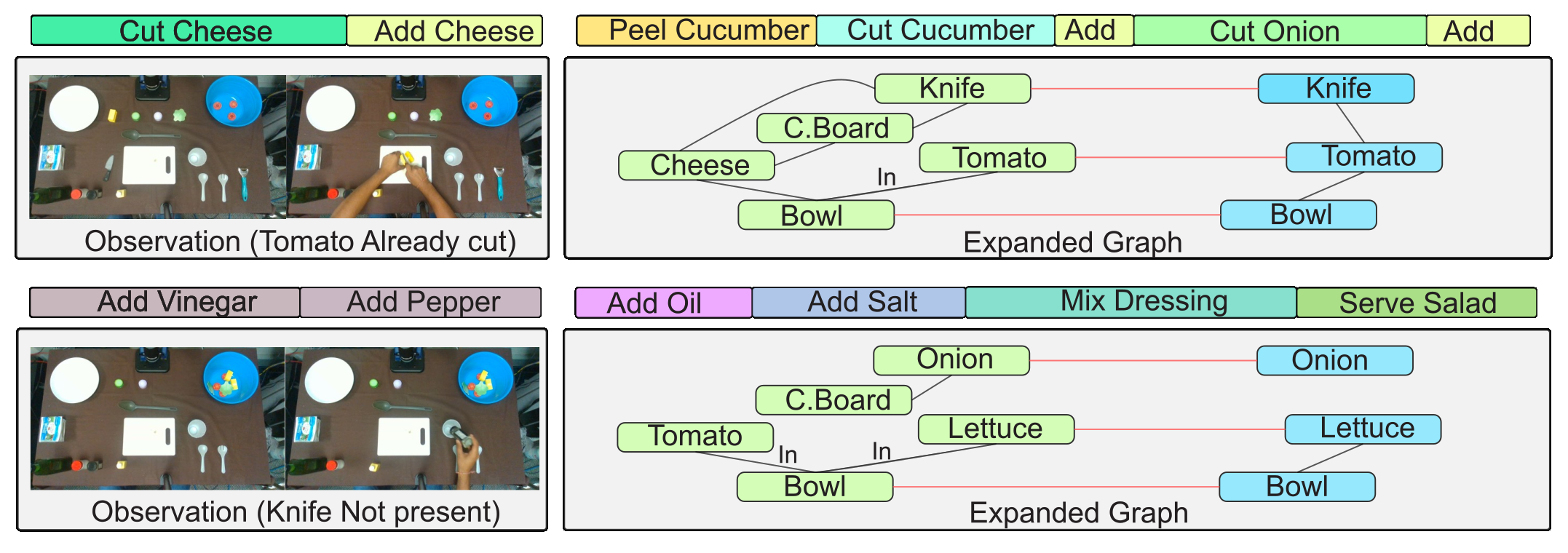}
\caption{Qualitative evaluation of scene and knowledge graph integration in a real-world kitchen scenario. 
Observed video and actions on the left, with predicted actions on the right, demonstrating the relevant portions of the expanded graphs (SG in green, KG in blue).
}
\label{tomato_already_cut}
\end{figure*}
\subsection{Baseline Comparison}
In~\cref{long_context_bf_50s}, we report anticipation performance with an extended observation context (\(\alpha = 0.2, 0.3\)), commonly used in prior works~\cite{gong2024actfusion, gong2022future, zhong2023diffant}. 
Our comparison includes methods that utilize action labels from observed video~\cite{abu2018will, abu2019uncertainty, ke2019time} as well as approaches that rely solely on latent features such as I3D~\cite{\droptext{}abu2021long\droptext{}, gong2022future, zhang2024object, patsch2024long, mate2024temporal}, and LLM baselines ranging from general-purpose~\cite{guo2025deepseek, openai2024o1} to domain-adapted~\cite{wang2025multimodal}. 
\droptext{}
\keeptext{Furthermore, we evaluate against recent alternative formulations: diffusion-based methods~\cite{zhong2023diffant, gong2024actfusion, zatsarynna2025manta} that use a denoising process akin to our iterative refinement, the stochastic memory-propagation approach MixANT~\cite{wasim2025mixant}, and S-GEAR~\cite{diko2024semantically}, which guides anticipation using implicitly distilled language-model semantics rather than an explicit graph.}
\droptext{}
\keeptext{Note that we exclude ANTICIPATR~\cite{nawhal2022rethinking} due to its different evaluation protocol~\cite{zhong2023survey}, as well as NeSCA~\cite{bhagat2023knowledgeguided}, which does not report results for this extended observation setting.}
SymAnt outperforms non-finetuned LLM baselines~\cite{guo2025deepseek,openai2024o1} by an average of $15\%$. 
While methods like ActionLLM~\cite{wang2025multimodal}, TCCA, \keeptext{S-GEAR,} and MixANT~\cite{wasim2025mixant} show strong performance at the shorter prediction horizon ($\beta=0.1$), SymAnt achieves the best MoC for all $\beta \geq 0.2$ on 50 Salads and on Breakfast at $\alpha=0.2$; on Breakfast at $\alpha=0.3$, TCCA and DiffAnt lead by at most $1.3$ MoC, within one standard deviation of our results. 
Specifically, on the 50 Salads dataset ($\alpha=0.2$), SymAnt outperforms the ActionLLM by over $8\%$ MoC and TCCA by over $6\%$ MoC at the longest horizon ($\beta=0.5$). 
These results, particularly for longer prediction horizons, validate the effectiveness of our approach in preserving long-term contextual information, a setting where other symbolic and non-symbolic baselines suffer from performance degradation.

We further present a comprehensive evaluation of our method on the EPIC-KITCHENS and EGTEA Gaze+ datasets in~\cref{tab:epic_egtea_params}, shifting the camera perspective from third-person to egocentric. 
Our comparison spans conventional methods~\cite{carreira2017quo, girdhar2017actionvlad, hussein2019timeception, hussein2019videograph, nagarajan2020ego}, dedicated transformer and diffusion architectures~\cite{nawhal2022rethinking, zhong2023diffant}, and the latest large-scale fine-tuned LLM-based models~\cite{zhao2023antgpt, kim2024palm, cao2025vision, chu2025intention}.
Notably, SymAnt achieves competitive performance with leading methods such as INSIGHT (8.5B) and ICVL (17.5B) while requiring only 3M parameters.
This underscores the effectiveness of bootstrapping symbolic contextual knowledge, enabling strong anticipation performance without relying on large-scale model capacity.
\keeptext{Furthermore, to test SymAnt beyond cooking, we evaluate on Assembly101~\cite{sener2022assembly101}, a large-scale dataset of toy-vehicle assembly and disassembly. Its knowledge graph is obtained automatically with the same procedure as in~\cref{sec:gsnn}, applied to the Assembly101 object vocabulary. We compare in~\cref{tab:assembly101} against transformer-based (FUTR~\cite{gong2022future}), diffusion-based (GTDA~\cite{zatsarynna2024gtda}, MANTA~\cite{zatsarynna2025manta}), and recent stochastic (MixANT~\cite{wasim2025mixant}) methods that report this benchmark. SymAnt outperforms the strongest baseline, MixANT, by $0.8$ MoC at $\alpha{=}0.2$, $\beta{=}0.5$ and by $0.5$ MoC at $\alpha{=}0.3$ for both $\beta{=}0.3$ and $\beta{=}0.5$. As on the cooking benchmarks, the advantage appears at the longest horizon, and the same graph-construction procedure transfers to this domain without domain-specific engineering.}

\subsection{Impact of Knowledge Editing}
\label{sec:kg_editing}
We conducted both qualitative and quantitative analyses to evaluate the effectiveness of incorporating and editing symbolic context from scene and knowledge graphs, demonstrating how modifications to the graph structure directly influence anticipation of future actions without retraining.

\begin{figure}[tbp]
    \centering
        \includegraphics[width=\linewidth]{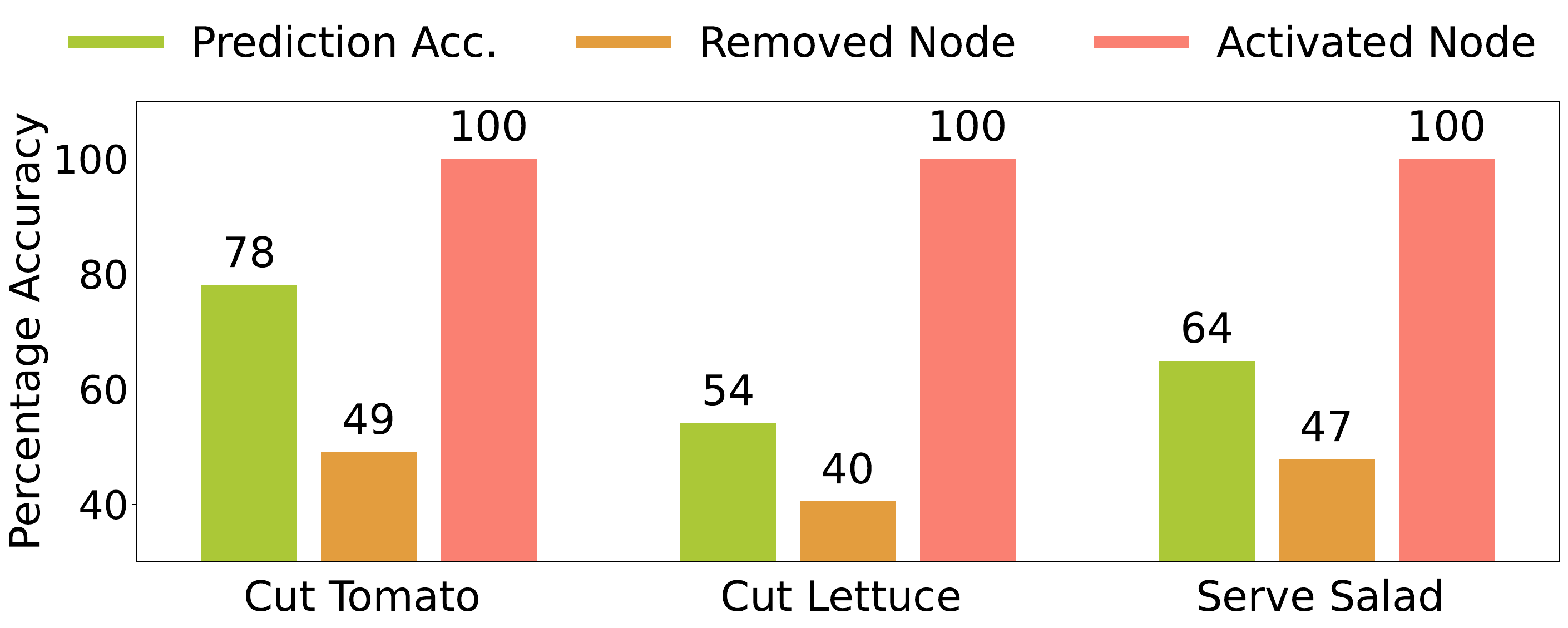}
        \caption{Quantitative analysis of anticipation with manual graph edits: accuracy drops on deactivation and is $100\%$ on activation.}
        \label{fig:symbolic_impact}
\end{figure}

To qualitatively demonstrate the impact of symbolic knowledge, we altered key components in the observation (e.g., removing a knife) which led to changes in the joint-graph, demonstrating symbolic information's effect on neural action prediction. 
We captured real-world videos in our dummy kitchen setup and
to adapt our model to this setup, we first fine-tuned it on the general salad-making activities dataset from~\cite{bhagat2023knowledgeguided}, recorded in a similar kitchen environment. 
We then evaluated our model on multiple cases with an observation horizon of $10\%$ ($\alpha=0.1$) to predict the subsequent $50\%$ ($\beta = 0.5$).
\Cref{tomato_already_cut} illustrates two such cases, along with their corresponding expanded joint-graphs (showing only the relevant parts) and the subsequently predicted actions.
In the first example, a tomato is already cut and placed in a bowl when the observation is made. The scene graph explicitly captures the ``in'' relation between the tomato and the bowl, allowing our model to recognize that no further tomato-related actions are necessary. %
In the second example, the scene graph correctly identifies the absence of a knife, and the knowledge graph further reasons about it, ensuring that no cutting action is predicted. 
Unlike purely neural models that might overlook these subtle contextual details, our method explicitly reasons over individual concepts in the scene.
Complementing the qualitative examples, we designed an experiment on the 50 Salads dataset to measure how activating or deactivating individual action nodes in the expanded graph influences the model's final predictions.
Specifically, we focus on the actions \textit{Cut Tomato}, \textit{Cut Lettuce}, and \textit{Serve Salads}. For each action, we generated a filtered subset in which the action is guaranteed to appear in the ground-truth anticipation. Anticipation accuracy for that action is defined as the percentage of samples in which the action appears in any of the model's $N$ predictions. We compare three conditions: (1) baseline accuracy without graph modifications, (2) manual deactivation of the target node, and (3) manual activation of the target node. The results, shown in~\cref{fig:symbolic_impact}, reveal accuracy variations corresponding to the different graph modifications. Removing the target node leads to a substantial drop in accuracy, whereas enforcing its activation increases accuracy to 100\%. 
These findings confirm that the symbolic initialization shapes the model's predictions, providing a retraining-free mechanism to steer anticipated actions.
\keeptext{Supplementary Sec.~D shows a realistic correction, where a node edit recovers an action missed due to an occluded object, and when the visual stream overrides edits.}
\begin{table}[tbp]
    \centering
        \caption{Performance (mAP) on EPIC-KITCHENS\keeptext{-55} and EGTEA Gaze+ across all, frequent, and rare splits. Parameter counts are approximated and report \emph{trainable} parameters only; for SymAnt, the frozen object detector, ViT node encoder, and I3D backbone are excluded. \droptext{}}
        \label{tab:epic_egtea_params}
        
        \resizebox{\linewidth}{!}{
        \begin{tabular}{l c ccc ccc} 
            \toprule
            Method & Params & \multicolumn{3}{c}{\textbf{EPIC-KITCHENS}} & \multicolumn{3}{c}{\textbf{EGTEA Gaze+}} \\
            \cmidrule(lr){3-5} \cmidrule(lr){6-8}
            & & All & Freq. & Rare & All & Freq. & Rare \\
            \midrule
            I3D \cite{carreira2017quo} & NA & 37.7 & 53.5 & 23.0 & 72.1 & 79.3 & 53.3 \\
            ActionVLAD \cite{girdhar2017actionvlad} & NA & 29.2 & 53.5 & 18.6 & 73.3 & 79.0 & 58.6 \\
            TimeCeption \cite{hussein2019timeception} & NA & 35.6 & 55.9 & 26.1 & 74.1 & 79.7 & 59.7 \\
            VideoGraph \cite{hussein2019videograph} & NA & 22.5 & 49.4 & 14.0 & 67.7 & 77.1 & 47.2 \\
            EGO-TOPO \cite{nagarajan2020ego} & NA & 38.0 & 56.9 & 29.2 & 73.5 & 80.7 & 54.7 \\
            ANTICIPATR \cite{nawhal2022rethinking} & NA & 39.1 & 58.1 & 29.1 & 76.8 & 83.3 & 55.1 \\
            DiffAnt \cite{zhong2023diffant} & 80M & 38.1 & 55.0 & 31.0 & 77.3 & 83.5 & 61.4 \\
            AntGPT \cite{zhao2023antgpt} & 7.5B & 40.1 & 58.8 & 31.9 & 80.2 & 84.8 & 72.9 \\
            PALM \cite{kim2024palm} & 7.5B & 40.4 & 59.3 & 30.3 & 80.7 & 85.0 & 73.5 \\
            Ours & 3M & 42.47 & 59.32 & 32.27 & 78.66 & 84.96 & 65.51 \\
            ICVL \cite{cao2025vision} & 17.5B & \underline{43.3} & \underline{61.6} & \underline{33.8} & \underline{81.0} & \underline{85.2} & \underline{73.7} \\
            INSIGHT \cite{chu2025intention} & 8.5B & \textbf{45.2} & \textbf{62.4} & \textbf{36.0} & \textbf{81.7} & \textbf{85.9} & \textbf{74.4} \\
            \bottomrule
        \end{tabular}
        }
\end{table}
\begin{table}[t]
    \centering
\caption{\keeptext{Long-term dense anticipation on Assembly101 (MoC).  
$^*$: retrained on Assembly101~\cite{zatsarynna2024gtda}.}
}
        \label{tab:assembly101}
        \footnotesize
        \resizebox{\linewidth}{!}{
        \begin{tabular}{l cccc cccc}
            \toprule
            & \multicolumn{4}{c}{$\beta$ ($\alpha=0.2$)} & \multicolumn{4}{c}{$\beta$ ($\alpha=0.3$)} \\
            \cmidrule(lr){2-5} \cmidrule(lr){6-9}
            Method & 0.1 & 0.2 & 0.3 & 0.5 & 0.1 & 0.2 & 0.3 & 0.5 \\
            \midrule
            Tri-gram~\cite{abu2019uncertainty} & 2.8 & 2.2 & 1.9 & 1.5 & 3.5 & 2.7 & 2.3 & 1.8 \\
            UAAA$^*$~\cite{abu2019uncertainty} & 2.7 & 2.1 & 1.8 & 1.6 & 2.4 & 2.1 & 1.9 & 1.7 \\
            FUTR$^*$~\cite{gong2022future} & \underline{7.5} & \underline{5.5} & 4.7 & 4.1 & \underline{7.8} & 6.0 & 5.2 & 4.0 \\
            GTDA~\cite{zatsarynna2024gtda} & 6.4 & 4.5 & 3.5 & 2.8 & 5.9 & 4.2 & 3.5 & 2.9 \\
            MANTA~\cite{zatsarynna2025manta} & 6.7 & 5.3 & 4.2 & 3.5 & 6.6 & 4.7 & 4.2 & 3.5 \\
            MixANT~\cite{wasim2025mixant} & \textbf{8.0} & \textbf{7.0} & \textbf{6.2} & \underline{4.4} & \textbf{8.7} & \textbf{6.6} & \underline{5.6} & \underline{4.6} \\
            Ours & 6.15 & 5.44 & \underline{5.39} & \textbf{5.16} & 7.15 & \underline{6.32} & \textbf{6.08} & \textbf{5.14} \\
            \bottomrule
        \end{tabular}
        }
\end{table}
\subsection{Ablation Studies}
In the following section, 
we ablate the individual components of our framework, and report inference-time analyses. 
\subsubsection{Encoder, Decoder and Knowledge Integration}
\label{enc_dec}
\label{sec:kg_ablation}
We conduct an ablation study to compare different choices for the encoder, decoder, and symbolic knowledge integration, and assess their impact on action anticipation performance. \Cref{tab:combined_encoder_decoder_bf_50s_alpha02} presents results on the Breakfast and 50 Salads datasets on $\alpha = 0.2$, omitting results for $\beta = 0.2$ and $0.3$ as they follow a uniform trend across different values of $\beta$. \keeptext{We also report standard deviations reflecting cross-split variation; since all configurations share these splits, consistent margins smaller than this spread remain meaningful. To test these differences, we conduct two-sided paired $t$-tests on the split-level averages.}
\begin{table}[tbp]
        \centering
        \caption{Ablation on model configurations. \textbf{Enc.}: M (Mamba), T (Transformer). \textbf{Dec.}: R (Iterative Refinement), D (Diffusion), S (DDIM sampling \cite{song2020denoising}). \textbf{KG} (symbolic context): I (init.\ from joint SG+KG search), C (cond.\ on joint SG+KG search), K (init.\ from KG-only search, scene graph disabled), \keeptext{O (init.\ from max-pooled CLIP embeddings, no graph), A (init.\ from a GAT, no graph search), G (init.\ from a Visual Genome KG~\cite{krishna2017visual}).}}
        \label{tab:combined_encoder_decoder_bf_50s_alpha02}
        \footnotesize
        \setlength{\tabcolsep}{3pt}
        \begin{tabular*}{\linewidth}{@{\extracolsep{\fill}} ccc cc cc @{}}
            \toprule
            & & & \multicolumn{2}{c}{\textbf{Breakfast}} & \multicolumn{2}{c}{\textbf{50 Salads}} \\
            \cmidrule(lr){4-5} \cmidrule(lr){6-7}
            & & & \multicolumn{2}{c}{$\beta$ ($\alpha=0.2$)} & \multicolumn{2}{c}{$\beta$ ($\alpha=0.2$)} \\
            \cmidrule(lr){4-5} \cmidrule(lr){6-7}  
            Enc. & Dec. & KG & 0.1 & 0.5 & 0.1 & 0.5 \\
            \midrule
            T & T & $-$ & 27.03\keeptext{${}_{\pm1.4}$} & 20.82\keeptext{${}_{\pm0.3}$} & 36.77\keeptext{${}_{\pm5.3}$} & 20.12\keeptext{${}_{\pm6.9}$} \\
            M & T & $-$ & 26.70\keeptext{${}_{\pm1.6}$} & 20.25\keeptext{${}_{\pm0.8}$} & 36.20\keeptext{${}_{\pm8.3}$} & 21.60\keeptext{${}_{\pm2.7}$} \\
            M & R & $-$ & 27.20\keeptext{${}_{\pm1.1}$} & 23.60\keeptext{${}_{\pm0.4}$} & 37.20\keeptext{${}_{\pm4.2}$} & \droptext{}\keeptext{27.70}\keeptext{${}_{\pm5.6}$}  \\
            $-$ & R & I & 17.81\keeptext{${}_{\pm1.6}$} & 11.86\keeptext{${}_{\pm0.7}$} & 25.20\keeptext{${}_{\pm7.3}$} & 12.41\keeptext{${}_{\pm5.6}$} \\
            M & S & I & 22.41\keeptext{${}_{\pm2.7}$} & 15.56\keeptext{${}_{\pm0.7}$} & 29.76\keeptext{${}_{\pm4.5}$} & 21.61\keeptext{${}_{\pm7.6}$} \\
            M & D & C & \droptext{}\keeptext{27.50}\keeptext{${}_{\pm0.5}$} & 24.08\keeptext{${}_{\pm0.5}$} & 38.70\keeptext{${}_{\pm6.2}$} & 27.44\keeptext{${}_{\pm6.4}$} \\
            M & R & C & 27.10\keeptext{${}_{\pm1.6}$} & \droptext{}\keeptext{25.41}\keeptext{${}_{\pm1.0}$} & 39.58\keeptext{${}_{\pm6.5}$} & 27.61\keeptext{${}_{\pm6.7}$} \\
            M & R & K & 26.67\keeptext{${}_{\pm2.0}$} & 24.56\keeptext{${}_{\pm0.5}$} & 38.88\keeptext{${}_{\pm1.6}$} & 27.44\keeptext{${}_{\pm6.5}$} \\
            \keeptext{M} & \keeptext{R} & \keeptext{O} & \keeptext{25.53${}_{\pm0.9}$} & \keeptext{19.61${}_{\pm0.5}$} & \keeptext{34.47${}_{\pm2.0}$} & \keeptext{21.58${}_{\pm3.5}$} \\
            \keeptext{M} & \keeptext{R} & \keeptext{A} & \keeptext{27.12${}_{\pm0.4}$} & \keeptext{24.54${}_{\pm0.3}$} & \keeptext{\underline{39.70}${}_{\pm7.7}$} & \keeptext{27.35${}_{\pm5.0}$} \\
            \keeptext{M} & \keeptext{R} & \keeptext{G} & \keeptext{\underline{27.85 ${}_{\pm0.9}$}} & \keeptext{\underline{25.55 ${}_{\pm0.6}$}} & \keeptext{38.50${}_{\pm3.8}$} & \keeptext{\underline{28.10 ${}_{\pm4.9}$}} \\
            M & R & I & \textbf{28.22}\keeptext{${}_{\pm1.8}$} & \textbf{25.67}\keeptext{${}_{\pm0.6}$} & \textbf{40.78}\keeptext{${}_{\pm5.3}$} & \textbf{28.19}\keeptext{${}_{\pm5.1}$} \\
            \bottomrule
        \end{tabular*}
\end{table}
Rows 1 and 2 \droptext{}\keeptext{compare} the Mamba (M) and Transformer (T) encoders\keeptext{. We do not detect a significant difference between them on either dataset ($p=0.126$ and $p=0.151$),} suggesting that Mamba is an efficient, drop-in replacement for Transformer given its faster inference speeds (~\cref{tab:inference_times}). Comparing rows 2 and 3 demonstrates that incorporating an iterative action decoder (Refinement, R) significantly enhances long-horizon action anticipation, particularly for larger $\beta$ values.
Additionally, rows 3 and 12 highlight the impact of knowledge-guided initialization (I), improving MoC in every setting, by $1.8\%$ on average.
Rows 4 and 12 compare the impact of symbolic initialization without video feature conditioning. The lower performance in row 4 underscores the importance of low-level visual features for accurate predictions. Comparing rows 5 and 12 highlights the impact of consistency between the training and inference steps for the iterative refinement process. Row 5 adopts a strategy inspired by DDIM~\cite{song2020denoising}, where the model is trained on a dense 1000 step refinement process but utilizes accelerated sampling with only 25 inference steps (S). This yields significantly lower accuracy than our proposed method.

Furthermore, rows 6 and 12 reveal that a traditional diffusion-based decoder Diffusion(D)~\cite{zhong2023diffant, gong2024actfusion}, which follows a conventional forward-reverse denoising process, underperforms our iterative refinement approach in every setting.
Finally, rows 7 and 12 show that conditioning the decoder on symbolic knowledge (C) instead of using it for initialization lowers accuracy in every setting, emphasizing the effectiveness of our design choice. Rows 8 and 12 further isolate the scene graph's contribution: a KG-only search (no SG) costs $1$--$2\%$ MoC.
\keeptext{Rows 9 and 10 replace the graph-search with competing knowledge-integration strategies. Max-pooled CLIP~\cite{radford2021clip} embeddings of the detected objects (O) fall below the no-context row 3 in every setting, showing that unstructured semantic input is not by itself beneficial. A 2-hop GAT~\cite{velickovic2018gat} (instead of graph-search network) over the same joint-graph nodes (A) recovers only part of the gain and stays $0.8$--$1.1\%$ MoC below row 12, highlighting the strength of our approach. Row 11 (G) uses the same architecture as row 12 but a Visual Genome KG~\cite{krishna2017visual} of kitchen objects plus one-hop neighbors ($714$ nodes vs.\ $175$ in I). At $\beta{=}0.5$, G scores $25.55$ (Breakfast) and $28.10$ (50 Salads), within $0.2$ MoC of I ($25.67$/$28.19$) and above the no-KG row ($23.60$/$27.70$). This shows that a dataset-specific knowledge graph is not a prerequisite: a general-purpose source already carries most of the long-horizon gain, without extra domain-specific engineering.}
Crucially, as detailed in the supplementary material, we observe identical performance dynamics on the EPIC-KITCHENS and EGTEA Gaze+.

\droptext{}

\subsubsection{Inference Speed and Efficiency}
\label{sec:inference_speed}
Beyond prediction accuracy, we assess the computational efficiency of our method. \cref{tab:inference_times} reports inference times (in ms) for different model configurations, computed on an NVIDIA RTX 6000 GPU. 
Comparing Mamba (M) and Transformer (T) encoders, Mamba reduces runtime from 1.5 to 1.1 ms without a significant drop in accuracy, owing to its fixed-size state matrix and optimized architecture. The Iterative Refinement (R) decoder scales linearly with the number of steps; we use 10 steps, yielding 14.5 ms while balancing accuracy and latency (see the supplementary material for a step-vs-accuracy analysis). The symbolic context aggregation module adds only 0.9 ms to the core pipeline runtime of 16.5 ms, demonstrating that the joint-graph search improves performance with minimal computational overhead. 
In a real-world streaming deployment, initial detections can be cached in a sliding window, so detector cost is a one-frame offset per prediction step (see~\cref{sec:gsnn}). Total per-step latency is ${\sim}21.5$ ms with YOLO, ${\sim}31.5$ ms with Faster R-CNN, and ${\sim}56.5$ ms with Grounding DINO. As the $\dagger$ row in~\cref{tab:inference_times} shows, swapping Grounding DINO for Faster R-CNN nearly halves latency while changing MoC by under $0.7\%$, enabling a speed-cost trade-off without retraining. This also demonstrates that our pipeline is robust to the noise of object detection. 
This efficiency also extends to parameter footprint: as shown by the parameter counts in~\cref{tab:epic_egtea_params} and the model-size analysis in the supplementary material, SymAnt maintains strong anticipation accuracy with far fewer parameters than most baselines, highlighting the advantages of our neuro-symbolic design.
\begin{table}[tbp]
    \centering
        \caption{Module-wise inference time (ms) on 50 Salads; all rows use KG initialization (I). Cont.: symbolic context aggregation. $\dagger$: SG built with Faster R-CNN.}
        \label{tab:inference_times}
        
        \setlength{\tabcolsep}{1.3pt}
        \footnotesize
        \begin{tabular*}{\linewidth}{@{\extracolsep{\fill}} cc ccc cccc @{}}
            \toprule
            & & \multicolumn{3}{c}{\textbf{Inference Time}} & \multicolumn{4}{c}{\textbf{Performance}} \\
            \cmidrule(lr){6-9} 
            & & \multicolumn{3}{c}{\textbf{(ms)}} & \multicolumn{2}{c}{$\beta$ ($\alpha=0.2$)} & \multicolumn{2}{c}{$\beta$ ($\alpha=0.3$)} \\
            \cmidrule(lr){3-5} \cmidrule(lr){6-7} \cmidrule(lr){8-9}
            Enc. & Dec. & Enc. & Dec. & Cont. & 0.1 & 0.5 & 0.1 & 0.5 \\
            \midrule
            T & T & 1.5 & 1.6 & 0.9 & 39.48 & 23.60 & 34.65 & 19.50 \\
            M & T & 1.1 & 1.6 & 0.9 & 39.16 & 23.84 & 34.72 & 19.46 \\
            M & R(10) & 1.1 & 14.5 & 0.9 & \textbf{40.78} & \textbf{28.19} & \underline{35.23} & \textbf{25.50} \\
            M & R(10)$^\dagger$ & 1.1 & 14.5 & 0.9 & 40.41 & 27.55 & 35.13 & 25.01 \\
            M & R(100) & 1.1 & 110 & 0.9 & \underline{40.56} & \underline{27.98} & \textbf{35.35} & \underline{25.25} \\
            \bottomrule
        \end{tabular*}
\end{table}
\section{Conclusion}

In this work, we propose SymAnt, a neuro-symbolic approach that unifies scene and knowledge graphs within a joint-graph search framework, providing an information-rich prior for long-term action anticipation.
In experiments across \droptext{}\keeptext{five} diverse benchmarks covering first- and third-person observations\keeptext{, cooking and assembly activities}, we show that this symbolic grounding enables accurate long-horizon anticipation, particularly from short observation contexts, while requiring significantly fewer parameters than most neural alternatives. 
Moreover, because SymAnt's predictions are anchored in explicit symbolic graph structures, users can activate or deactivate graph nodes at inference time to steer anticipated actions. These properties position SymAnt as a step towards action anticipation systems that are both data-efficient and adaptable to changing environments and task specifications. We discuss limitations and future work in the supplementary material.

 {
    \small
    \bibliographystyle{ieeenat_fullname}
    \bibliography{main}
}

\maketitlesupplementary
\appendix
\section{Methodology Details}
\subsection{Algorithms} 
In this section, we provide detailed algorithm descriptions for the components of our method. First,~\cref{alg:graph_expansion} describes our overall joint graph-search network.~\cref{alg:temporal_sg_generation_symbols_updated_v6} describes the process of generating the temporal scene graph. ~\cref{alg:graph_merge} describes our graph-merging algorithm and~\cref{alg:temporal_embedding} describes our iterative action denoising approach.

\subsection{Training Losses}
\label{sec:losses}
This section details the loss functions used in end-to-end training SymAnt. We begin with the action recognition loss for observed actions, followed by the importance loss for joint-graph search. Finally, we describe the iterative refinement loss and initialization loss for the decoder.

\subsubsection{Recognition Loss} 

We apply an action recognition loss to align the predicted and ground truth action sequences for the observed portion of the video. This serves a dual purpose: first, it ensures that the model learns to recognize past actions accurately; second, it enhances the quality of past video embeddings, making them more informative for future action anticipation. 
The recognition loss \( \mathcal{L}^{\text{rec}} \) is formulated as a cross-entropy loss between the predicted action logits  ${\mathbf{A}}^{\text{obs}}$ and ground truth action $\tilde{\mathbf{A}}^{\text{obs}}$ as:

 \begin{equation}
    \mathcal{L}^{\text{rec}} = -\sum_{i=1}^{L} \sum_{j=1}^{C} \mathbf{A}^{\text{obs}}_{i,j} \log \tilde{\mathbf{A}}^{\text{obs}}_{i,j} 
 \end{equation}

where $L$ is the observed video frames and $C$ is the number of actions.  

\subsubsection{Importance Loss}
The importance loss ensures that the joint-graph search network correctly identifies and activates the relevant actions and affordances. To achieve this, we compute $L2$ loss between the predicted importance scores \( \tilde{\mathbf{I}} \) and the ground truth importance scores \( \mathbf{I} \):

\begin{equation}
  \mathcal{L}^{\text{imp}} = \sum_{w=1}^{W} {(\mathbf{I}^{w} - \tilde{\mathbf{I}}^{w}) }^2
\end{equation}

where $W$ represents the set of active nodes along with their directly connected inactive neighbors. The ground truth importance scores are derived from the future action labels. Specifically, nodes corresponding to actions, affordances, and entities associated with the ground-truth future actions are assigned an importance score of \( 1 \), while all other nodes are assigned a score of \( 0 \). This loss guides the model to focus on the most relevant parts of the joint-graph.

\subsubsection{Action Refinement Loss}

The action refinement loss $\mathcal{L}^{\text{AR}}$ ensures both the proper functioning of the decoder’s iterative refinement steps and the accuracy of the final predicted actions. It encourages the decoder to progressively refine its predictions while ensuring that the final output actions remain reliable. This loss is computed as a weighted sum of the anticipation loss (defined later) at the intermediate refinement steps and at the final output. The action refinement loss is given by:

\begin{equation}
 \mathcal{L}^{\text{AR}} =  \lambda\sum_{t=T-1}^{1} \mathcal{L}^{\text{ant}}(\tilde{\mathbf{A}_t}, \tilde{\mathbf{d}_t}) + \mathcal{L}^{\text{ant}}(\tilde{\mathbf{A}}, \tilde{\mathbf{d}})    
\end{equation}

where \(T\) is the number of steps in the denoising process, and \(\lambda\) is the weight for the intermediate losses, taken as \(1/T\).

The anticipation loss is a combination of action loss and duration loss. 
The action loss $\mathcal{L}^\text{ant}_{\text{action}}$ is defined as the cross-entropy between ground-truth actions $\mathbf{A}$, which are provided by the dataset and predicted actions $\tilde{\mathbf{A}}$, while the duration regression loss $\mathcal{L}^{\text{ant}}_{\text{duration}}$ is defined using the $L2$ loss between target durations $\mathbf{d}$ and predicted durations $\tilde{\mathbf{d}}$. 

\begin{equation}
    \mathcal{L}^\text{ant}_{\text{action}}= - \sum_{i=1}^{N} \sum_{j=1}^{C} \mathbf{A}_{i,j} \log (\tilde{\mathbf{A}}_{i,j})
\end{equation}

\begin{equation}
    \mathcal{L}^{\text{ant}}_{\text{duration}} = \sum_{i=1}^{N} (\mathbf{d}_{i} - \tilde{\mathbf{d}}_i)^2
\end{equation}

\subsubsection{Initialization Loss:} The initialization loss ensures that the symbolic initialization is correctly interpreted by the neural decoder and that this information is preserved throughout the iterative denoising process. Specifically, it enforces that action nodes that are active in the fully expanded joint-graph remain present in the final predictions. 
The initialization loss is formulated as cross-entropy between active action nodes in the expanded joint-graph and the predicted action logits $\tilde{\mathbf{A}}$: 
\begin{equation}
    \mathcal{L}^{\text{init}} =  - \sum_{i=1}^{N} \sum_{j=1}^C \mathbbm{1}_{j \in C_K} \log(\tilde{\mathbf{A}}_{i,j})
\end{equation}

where \( \mathbbm{1}_{j \in \mathbf{C}_K} \) is an indicator function that is 1 if \( j \in \mathbf{C}_K \) and $\mathbf{C}_K$ is a subset of actions that are active in the joint-graph after expansion.

\subsubsection{Total Loss}

The total training loss \(\mathcal{L}_{\text{total}}\) for SymAnt is a weighted summation of all individual loss components described above. These include the recognition loss, importance loss, action refinement loss (which itself includes anticipation and duration losses), and the initialization loss. Formally, it is defined as:

\begin{equation}
    \mathcal{L}_{\text{total}} = \mathcal{L}^{\text{rec}} +  \gamma \mathcal{L}^{\text{imp}} +\mathcal{L}^{\text{AR}} + \delta \mathcal{L}^{\text{init}}
\end{equation}

where \(\gamma\), and \(\delta\) are scalar weights that control the contribution of importance and initialization loss terms during training which are typically taken 0.1 and 0.01 respectively. 

\begin{algorithm}[t]
\caption{Joint-Graph Search Network}
\label{alg:graph_expansion}
\begin{algorithmic}

\STATE \textbf{Input:} Observed Video Features: $\mathbf{F}$ ;  Active Nodes : $\mathbf{M}_{active}$ 

\WHILE{True}
    \STATE \textbf{Step 1:} Get all inactive neighbors of $\mathbf{M}_{active}$
    \STATE $\mathbf{M}_{nei}$ $\gets$ get\_neighbors($\mathbf{M}_{active}$, $\mathbf{M}$
)

    \STATE \textbf{Step 2:} Propagate active nodes and neighbors 
    \STATE $\mathbf{M}_{nei}$, $\mathbf{M}_{active}$ $\gets$ $f_{prop}(\mathbf{M}_{nei}$, $\mathbf{M}_{active}$, $\mathbf{M}$)

    \STATE \textbf{Step 3:} Identify Important Neighbors
    \FOR{each $\mathbf{M}_{nei}^i \in \mathbf{M}_{nei}$}
        \STATE $\mathbf{I}^i \gets$ $f_{imp}(\mathbf{M}_{nei}^i$, $\mathbf{F}$)  
        \IF{$\mathbf{I}^i > \tau$}
            \STATE $\mathbf{M}_{active}$ $\gets$ $\mathbf{M}_{active}$ + $\mathbf{M}_{nei}^i$
        \ENDIF
    \ENDFOR

\STATE \textbf{Step 4:} Dynamic Propagation for Expansion
\STATE $\mathbf{I}^{\text{max}} \gets \max(\mathbf{I})$

\IF{ $\mathbf{I}^{\text{max}} < \eta$}
    \STATE \textbf{Break the loop:} Expansion threshold did not meet
\ENDIF
\ENDWHILE
\STATE \textbf{Step 5:} Get the context from fully expanded graph
\STATE $\mathbf{z}_T = f_{cont}(\mathbf{M}_{active}, \mathbf{F})$

\end{algorithmic}
\end{algorithm}

\begin{algorithm}[t]
\caption{Temporal Scene Graph Generation}
\label{alg:temporal_sg_generation_symbols_updated_v6}
\begin{algorithmic}

\STATE \textbf{Input:} Observed video $V_o = \{V_o^1, \dots, V_o^L\}$; Edge decay factor $\kappa$
\STATE \textbf{Initialize:} Node set $\mathbf{S}_n \gets \emptyset$, Edge set $\mathbf{S}_A \gets \emptyset$

\FOR{$l = 1$ to $L$}
    \STATE \textbf{Step 1:} Detect concepts in the current frame
    \STATE $\mathbf{c}^l \gets \text{GroundedSAM}(V_o^l)$
    \FOR{each concept $c_i \in \mathbf{c}^l$}
        \IF{$c_i \notin \mathbf{S}_n$}
            \STATE $\mathbf{S}_n \gets \mathbf{S}_n \cup \{c_i\}$
        \ENDIF
    \ENDFOR

    \STATE \textbf{Step 2:} Update temporal edges
    \FOR{each concept pair $(c_i, c_j) \in \mathbf{c}^l$}
        \STATE $(r_{ij}, w_{ij}, \text{exists}_{ij}) \gets f_{\text{edge}}(c_i, c_j)$
        
        \IF{$\text{exists}_{ij} = \text{True}$}
            \IF{$\exists e_{ij} = (c_i, c_j, r', w') \in \mathbf{S}_A$} 
                \IF{$r' \neq r_{ij}$}
                    \STATE Decay previous edge: $w' \gets w' \cdot \kappa$
                    \STATE Add updated edge: $e_{\text{new}} \gets (c_i, c_j, r_{ij}, w_{ij})$
                    \STATE $\mathbf{S}_A \gets \mathbf{S}_A \cup \{e_{\text{new}}\}$
                \ENDIF
            \ELSE
                \STATE Add new edge: $e_{\text{new}} \gets (c_i, c_j, r_{ij}, w_{ij})$
                \STATE $\mathbf{S}_A \gets \mathbf{S}_A \cup \{e_{\text{new}}\}$
            \ENDIF
        \ENDIF

    \ENDFOR
\ENDFOR

\STATE \textbf{Output:} Temporal Scene Graph $\mathbf{S} = (\mathbf{S}_n, \mathbf{S}_A)$

\end{algorithmic}
\end{algorithm}

\begin{algorithm}[t]
\caption{Graph Merging Process}
\label{alg:graph_merge}
\begin{algorithmic}
\STATE \textbf{Input:} Scene Graph $\mathbf{S} = \{\mathbf{S}_n, \mathbf{S}_A\}$; Knowledge Graph $\mathbf{K} = \{\mathbf{K}_n, \mathbf{K}_A\}$

\STATE \textbf{Initialize}: Merged Graph $\mathbf{M} = \{\mathbf{M}_n, \mathbf{M}_A\} = (\mathbf{S}, \mathbf{K})$; Active Nodes $\mathbf{M}_{active} = \{\}$

\FOR{each $\mathbf{S}_n^i$ $\in$ $\mathbf{S}_n$}
    \STATE Connect $\mathbf{S}_n^i$ and $\mathbf{K}_n^j$ in $\mathbf{M}$, where $\mathbf{K}_n^j$ is the node corresponding to $\mathbf{S}_n^i$ in $\mathbf{K}$
    \STATE  $\mathbf{M}_{active}$ $\gets$ $\mathbf{M}_{active}$ + $\{\mathbf{S}_n^i, \mathbf{K}_n^j\}$
\ENDFOR

\end{algorithmic}
\end{algorithm}

\begin{algorithm}[t]
\caption{Iterative Symbolic Action Decoding}
\label{alg:temporal_embedding}
\begin{algorithmic}

\STATE \textbf{Input:} Symbolic Context: $\mathbf{z}_T$, Encoded Observations: $\mathbf{E}$, Action Queries: $\mathbf{Q}$
\STATE \textbf{Initialize:} $\mathcal{L}^{\text{AR}} \gets 0$ ; ${\mathbf{z}_T} \gets \mathbf{z}_T + \mathbf{Q}$
\FOR{$t = T$ to $1$}
    \STATE \textbf{Step 1:} Temporal Encoding
    \STATE 
    $\tilde{\mathbf{z}_t} \gets \mathbf{z}_t + \mathbf{T}_E(t)$
    \STATE \textbf{Step 2:} Refinement Step
    \STATE $\mathbf{z}_{t-1}$ = refinement\_step($\tilde{\mathbf{z}_t}$, $\mathbf{E}$)
    
    \STATE \textbf{Step 3:} Projection
    \STATE $\tilde{\mathbf{A}_{t-1}}, \tilde{\mathbf{d}_{t-1}}$ $\gets$ projection\_net($\mathbf{z}_{t-1}$)

    \STATE \textbf{Step 4:} Calculate Intermediate anticipation Loss
    \STATE $\mathcal{L}^{\text{AR}} \gets \mathcal{L}^{\text{AR}} + \lambda \mathcal{L}^{\text{ant}}(\tilde{\mathbf{A}_{t-1}}, \tilde{\mathbf{d}_{t-1}})$

\ENDFOR

\STATE \textbf{Step 5:} Calculate Final Anticipation Loss
\STATE $\mathcal{L}^{\text{AR}} \gets  \mathcal{L}^{\text{AR}} +  \mathcal{L}^{\text{ant}}(\tilde{\mathbf{A}}, \tilde{\mathbf{d}})$

\end{algorithmic}
\end{algorithm}

\section{Additional Experimental Details}
\subsection{Datasets}
\label{sec:datasets}
\textbf{Breakfast}~\cite{kuehne2014language} and \textbf{50 Salads}~\cite{stein2013combining} are fixed camera datasets.
The Breakfast~\cite{kuehne2014language} dataset consists of 1,712 videos featuring 52 different individuals making breakfast in 18 different kitchens, totaling 77 hours of footage.
Each video is categorized into one of 10 activities related to breakfast preparation and annotated with 48 fine-grained actions. 
The 50 Salads~\cite{stein2013combining} dataset includes 50 top-view videos of 25 people preparing a salad, containing over 4 hours of RGB-D video data annotated with 17 fine-grained action labels and 3 high-level activities.  \textbf{Epic Kitchens}~\cite{damen2020epic} and \textbf{EGTEA Gaze+}~\cite{li2018eye} are egocentric datasets. 
Epic Kitchens comprises 39,596 segments labeled with 125 verbs, 352 nouns, and 2,513 verb-noun combinations (actions), totaling 55 hours of video. 
EGTEA Gaze+ contains 28 hours of videos with 10.3K action annotations, including 19 verbs, 51 nouns, and 106 unique actions.
\keeptext{\textbf{Assembly101}~\cite{sener2022assembly101} moves outside the cooking domain: it contains 362 assembly--disassembly sequences of 101 take-apart toy vehicles, recorded from 12 viewpoints for a total of 4{,}321 videos and $513$ hours, and we use its $202$ coarse action classes.}
We also utilize a \textbf{real-world mini-kitchen} setup from~\cite{bhagat2023knowledgeguided} for fine-tuning our model, which contains 20 videos recorded in a controlled kitchen environment.

\subsection{Training and Architecture Details}
\label{sec:parameters}

We configure the hidden dimension \( D \) to 128 for the Breakfast dataset and 512 for all other datasets. The number of action queries \( N \) is set to 8 for Breakfast and 20 for 50 Salads  \keeptext{and Assembly101}. For Epic Kitchens and EGTEA Gaze+, even though the task involves multi-label action classification~\cite{nagarajan2020ego}, we still predict sequential actions, setting \( N = 50 \). To ensure a fair comparison with state-of-the-art long-term action anticipation methods, we use pre-extracted I3D features~\cite{carreira2017quo} as input visual features \( F \) for \droptext{}\keeptext{the four cooking datasets}, provided by~\cite{farha2019ms} and~\cite{nagarajan2020ego}. \keeptext{For Assembly101 we instead use the TSM features released with the dataset~\cite{sener2022assembly101}, as done by the baselines on that benchmark.} We sample the I3D features with a stride of 3 for the Breakfast and 50 Salads datasets\keeptext{, the TSM features with the same stride for Assembly101,} and with a stride of 1 for Epic Kitchens and EGTEA Gaze+. During training, the observation rate \( \alpha \) is set to \{0.05, 0.1, 0.2, 0.3, 0.4, 0.5\} for Breakfast\droptext{}\keeptext{,} 50 Salads\keeptext{ and Assembly101}, and additionally to \{0.6, 0.7, 0.8\} for Epic Kitchens and EGTEA. We used the Adam optimizer and trained for 70 epochs for 50 Salads and 50 epochs for the other datasets. The total number of denoising steps is set to \( T = 10 \), and these configurations were optimized via hyperparameter tuning. The average training time across \droptext{}\keeptext{the four cooking datasets} is approximately 5 hours on an Nvidia RTX 6000 GPU, with an inference time of around 16.5 ms per video (using a pre-computed scene graph). Despite incorporating a graph-based symbolic network alongside traditional neural networks, training and inference times remain largely unchanged, as we generate only a single joint graph for each input video. During training, GPU utilization is approximately 18 GB.

\subsection{Large language models evaluation}
\label{llm_eval}
To compare our model’s performance with state-of-the-art multi-modal foundation models, we evaluated three reasoning models: ChatGPT-o1, Qwen3-Omni and Deepseek-r1. Specifically, for ChatGPT-o1 and Qwen3-Omni which can ingest any number of images before making predictions we sampled images from observed videos (based on the observation ratio $\alpha$) and prompted the model to predict subsequent future actions along with their relative durations. We then compared the predicted actions with the ground truth using a prediction ratio $\beta$, similar to our model’s evaluation. For Deepseek-r1, since its API does not accept images, we prompted it with the ground truth observed actions while keeping the remaining prompt and evaluation the same. Due to budget constraints, we evaluated ChatGPT-o1 using the 0.05 observation split of the Breakfast dataset; this split was chosen because it yields results closest to those on the 50 Salads dataset when compared to our model.

\subsection{Evaluation Metrics}
\textbf{Mean Accuracy over Classes (MoC)}: 
We evaluate performance on the Breakfast and 50 Salads datasets using the mean accuracy over classes (MoC). For a given anticipation duration, MoC is computed as the average accuracy per class over all future timestamps. Building on prior work~\cite{abu2018will,gong2022future,zhong2023diffant}, we first observe an initial portion of the video, denoted by \(\alpha\) (typically 0.2 or 0.3). To evaluate short-context action anticipation, we also include \(\alpha = 0.05\) and 0.1, following~\cite{bhagat2023knowledgeguided}. Anticipation begins immediately after the observed frames, but the predicted segment length is determined by \(\beta \in \{0.1,0.2,0.3,0.5\}\), which is defined as a fraction of the entire video—not just the unobserved portion. We evaluate performance across 4 splits for the Breakfast dataset and 5 splits for the 50 Salads dataset.

\textbf{Mean Average Precision (mAP)}: For the Epic Kitchens and EGTEA Gaze+ datasets, we evaluate performance using mean average precision (mAP) following~\cite{nagarajan2020ego}, which is a multi-label classification metric that quantifies the accuracy of predicting specific action classes. In our approach, an initial portion of each untrimmed video, \(\alpha\) is used as input to forecast all subsequent action classes, which occur over the remaining \((1-\alpha)\) duration of the video. Consistent with~\cite{nagarajan2020ego}, we set \(\alpha = \{0.25, 0.50, 0.75\}\) during evaluation and report mAP scores separately for low-shot (rare) and many-shot (frequent) scenarios.

\section{Further Ablations}
\subsection{Encoder, Decoder and Knowledge Integration}

We evaluate these encoder, decoder, and symbolic knowledge integration strategies on the EPIC-Kitchens and EGTEA Gaze+ datasets, as summarized in \cref{tab:combined_encoder_decoder_gtea}. Consistent with our prior observations, \droptext{} the Mamba (M) and Transformer (T) encoders \keeptext{perform comparably} (rows 1 and 2). However, the choice of decoder plays a critical role; incorporating an iterative denoising decoder (Refinement, R) substantially improves mAP (rows 2 and 4). Additionally, knowledge-guided initialization (I) yields an average mAP increase of 2.9\% across all splits (rows 4 and 5). Moreover, a conventional diffusion-based decoder (Diffusion, D) following a standard forward-reverse process significantly underperforms compared to our iterative refinement approach (rows 3 and 5). Ultimately, this confirms that explicit symbolic knowledge and our tailored iterative denoising effectively enhance performance over standard neural approaches across diverse datasets and observation ratios.

\begin{table}[t]
\centering
\caption{Comparison of anticipation accuracy across different model configurations for EPIC Kitchens and EGTEA Gaze+ dataset. Enc. (Encoder): M (Mamba), T (Transformer). Dec. (Decoder): R (Iterative Refinement), T (Transformer), D (Diffusion model with explicit forward process \cite{zhong2023diffant,gong2024actfusion}). KG (Knowledge Graph Integration): I (Initialization), C (Conditioning), $-$ (None).}
\label{tab:combined_encoder_decoder_gtea}

\footnotesize 

\resizebox{\linewidth}{!}{
\begin{tabular}{ccc ccc ccc}
    \toprule
    & & & \multicolumn{3}{c}{\textbf{EPIC Kitchens}} & \multicolumn{3}{c}{\textbf{EGTEA Gaze+}} \\
    \cmidrule(lr){4-6} \cmidrule(lr){7-9}
    Enc. & Dec. & KG & All & Frequent & Rare & All & Frequent & Rare \\
    \midrule
    T & T & $-$ & 37.9 & 55.3 & 29.3 & 72.1 & 75.4 & 59.2 \\
    M & T & $-$ & 38.6 & 56.1 & 29.7 & 72.1 & 76.8 & 59.4 \\
    M & D & C & 38.5 & 55.3 & 29.8 & 73.6 & 78.4 & 61.6 \\
    M & R & $-$ & \underline{39.7} & \underline{56.4} & \underline{31.8} & \underline{74.4} & \underline{79.7} & \underline{63.8} \\
    M & R & I & \textbf{42.4} & \textbf{59.3} & \textbf{32.2} & \textbf{78.7} & \textbf{84.9} & \textbf{65.5} \\
    \bottomrule
\end{tabular}
}
\end{table}

\subsection{Analysis of Refinement Steps}

\begin{figure}[t]
    \centering
    \includegraphics[width=\linewidth]{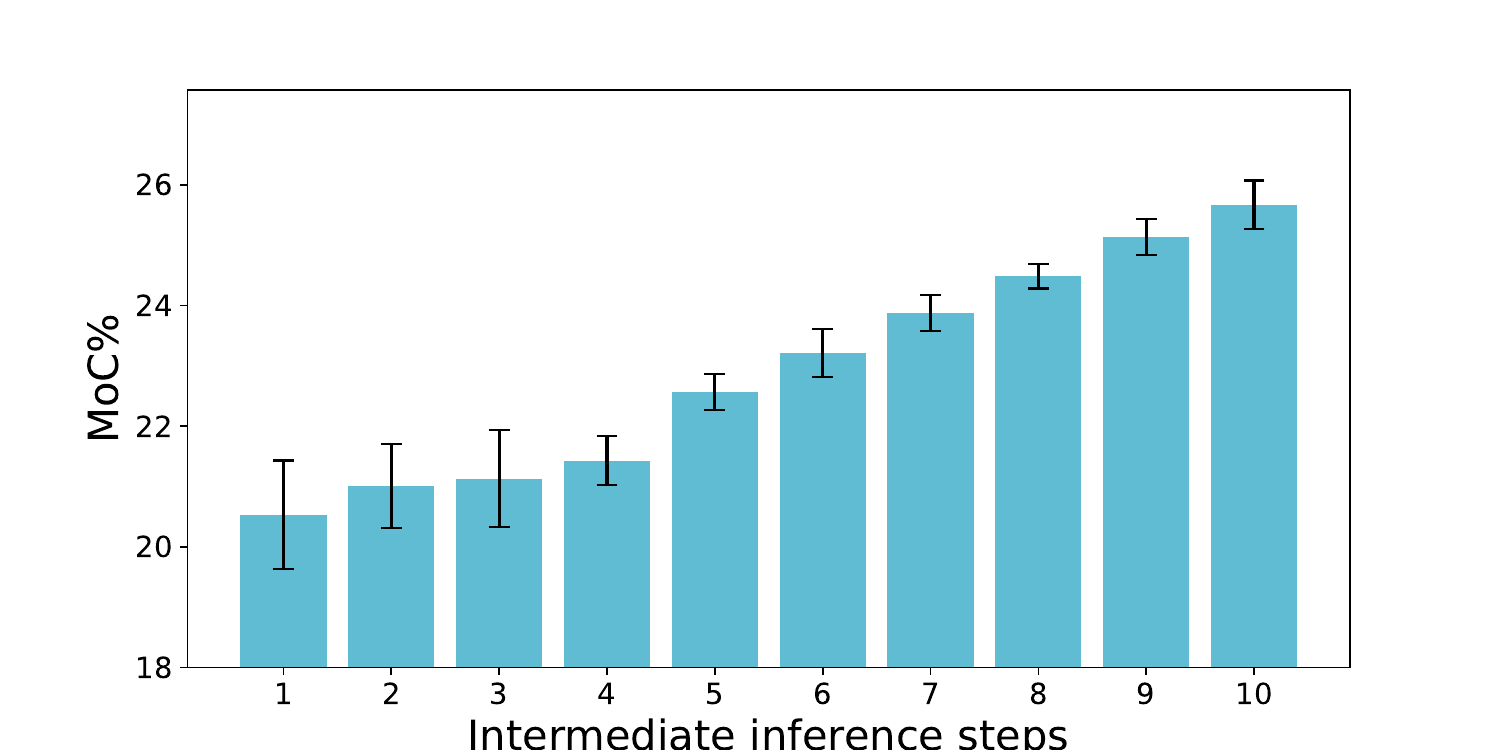}
    \caption{Analysis of intermediate refinement steps. We report mean accuracy and std. deviation over intermediate inference steps for a model trained for 10 steps.}
    \label{fig:intedmed_analysis}
\end{figure}

\begin{figure}[t]
    \centering
    \includegraphics[width=\linewidth]{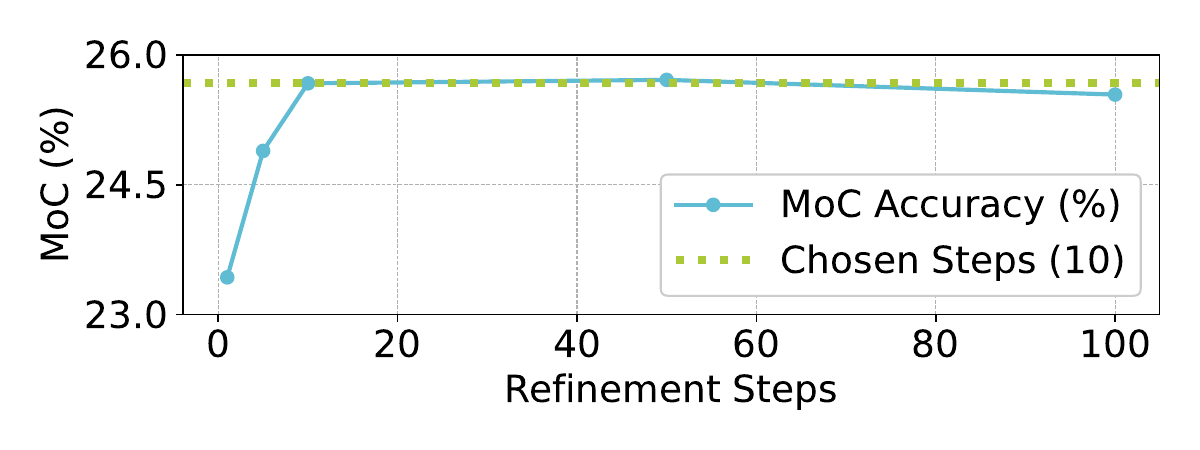}
    \caption{MoC accuracy vs. number of steps in the refinement process on the Breakfast dataset, showing that accuracy at 50 and 100 steps does not significantly improve compared to 10 steps.}
    \label{fig:stes_vs_accuracy}
\end{figure}

To analyze the impact of the number of refinement steps on model accuracy, we evaluated performance using different numbers of steps in the refinement process, as shown in~\cref{fig:stes_vs_accuracy}. In this experiment, we used Breakfast dataset with an observation ratio of 0.2 and a prediction ratio of 0.5. Since the accuracy achieved at 50 and 100 steps was not statistically better than the mean accuracy obtained at 10 steps, we selected 10 as the ideal number of steps to balance runtime and accuracy. Furthermore,~\cref{fig:intedmed_analysis} presents the accuracy at various intermediate refinement steps for a model trained with 10 refinement steps on the Breakfast dataset. For each split, we trained ten models and reported the overall mean and standard deviation of the accuracy at each inference step. Our observations reveal that the accuracy starts low, increases to a peak as more refinement steps are applied, and is accompanied by a decreasing standard deviation. This behavior indicates that our model is effectively capable of iteratively refining its anticipated actions.

\subsection{Impact of Different Loss Terms}

To evaluate the impact of different loss terms on the model's performance, we conducted an ablation study on various loss components using the Breakfast and 50 Salads datasets, as shown in~\cref{tab:combined_loss}. The results reveal that recognition loss significantly enhances accuracy, highlighting its importance in the model's performance. While initialization loss does not lead to substantial improvements in accuracy, both qualitative and quantitative analyses of symbolic knowledge's influence on the model's predictions suggest that it plays a critical role in maintaining proper alignment between the neural and symbolic components. This alignment, as discussed in the paper, ultimately improves the overall observability and coherence of the model's predictions.

\begin{table*}[t]
\centering
\caption{Results for different loss terms on the Breakfast and 50 Salads datasets, highlighting the significant impact of the recognition loss on model performance.}
\label{tab:combined_loss}

\footnotesize 

\begin{tabular*}{\textwidth}{@{\extracolsep{\fill}} ccc cccc cccc @{}}
\toprule
& & & \multicolumn{4}{c}{\textbf{Breakfast}} & \multicolumn{4}{c}{\textbf{50 Salads}} \\
\cmidrule(lr){4-7} \cmidrule(lr){8-11}
& & & \multicolumn{2}{c}{$\beta$ ($\alpha=0.2$)} & \multicolumn{2}{c}{$\beta$ ($\alpha=0.3$)} & \multicolumn{2}{c}{$\beta$ ($\alpha=0.2$)} & \multicolumn{2}{c}{$\beta$ ($\alpha=0.3$)} \\
\cmidrule(lr){4-5} \cmidrule(lr){6-7} \cmidrule(lr){8-9} \cmidrule(lr){10-11}   
AR & Rec. & Init. & 0.1 & 0.5 & 0.1 & 0.5 & 0.1 & 0.5 & 0.1 & 0.5 \\
\midrule
\checkmark & & & 23.75 & 19.21 & 26.64 & 23.28 & 32.19 & 22.05 & 30.61 & 17.52 \\
\checkmark & & \checkmark & 24.02 & 19.44 & 26.21 & 23.75 & 32.78 & 22.70 & 31.74 & 17.26 \\
\checkmark & \checkmark & & \underline{28.07} & \textbf{25.81} & \underline{34.12} & \underline{29.51} & \underline{40.69} & \underline{27.98} & \textbf{35.51} & \underline{25.31} \\
\checkmark & \checkmark & \checkmark & \textbf{28.22} & \underline{25.67} & \textbf{34.22} & \textbf{30.11} & \textbf{40.78} & \textbf{28.19} & \underline{35.23} & \textbf{25.50} \\
\bottomrule
\end{tabular*}
\end{table*}

\subsection{Parameter Efficiency}

\Cref{tab:model_performance} summarizes results on the Breakfast dataset for different model sizes. Our 3M parameter variant significantly outperforms both traditional baselines such as Diffant (80M) and fine-tuned LLM-based approaches such as LLMAction and ActionLLM (7B each), particularly at longer prediction horizons. Even our 400k variant retains high predictive accuracy, dropping only ${\sim}1.5\%$ compared to the baselines. As also reflected in the cross-dataset parameter counts reported in the main paper, this parameter efficiency generalizes across datasets, underscoring a central advantage of our neuro-symbolic approach: by bootstrapping symbolic knowledge as structured context, SymAnt achieves strong anticipation performance without relying on large-scale model capacity or extensive observation windows.

\begin{table}[t]
    \centering
        \caption{Parameter efficiency on the Breakfast dataset across different model size variants, showing high performance even with $<1M$ parameters. Counts are \emph{trainable} parameters of the anticipation model; the frozen I3D backbone (shared across all baselines) and the swappable SG detector (see the main paper) are excluded as common input pipeline.}
        \label{tab:model_performance}
        
        \footnotesize
        \begin{tabular*}{\linewidth}{@{\extracolsep{\fill}} l c cc cc @{}}
            \toprule
            & & \multicolumn{2}{c}{$\beta$ ($\alpha=0.2$)} & \multicolumn{2}{c}{$\beta$ ($\alpha=0.3$)} \\
            \cmidrule(lr){3-4} \cmidrule(lr){5-6}
            Method & Params & 0.2 & 0.5 & 0.2 & 0.5 \\
            \midrule
            Ours & 400k & 23.88 & 22.87 & 28.60 & 25.00 \\
            Ours & 800k & 24.54 & 21.58 & 30.61 & 27.63 \\
            FUTR & 1M & 24.55 & 22.04 & 29.88 & 25.87 \\
            Ours & 1.2M & \underline{27.54} & \underline{25.05} & 31.56 & 29.78 \\
            Ours & 3M & \textbf{27.55} & \textbf{25.67} & \textbf{32.06} & \underline{30.11} \\
            ActFusion & 13M & 25.52 & 23.25 & 31.76 & 28.78 \\
            Diffant & 80M & 24.59 & 22.74 & \underline{31.83} & \textbf{30.77} \\
            LLMAction & 7B & 25.23 & 24.26 & 31.35 & 27.81 \\
            ActionLLM & 7B & 23.32 & 20.43 & 29.45 & 24.55 \\
            \bottomrule
        \end{tabular*}
\end{table}

\section{\keeptext{Qualitative Results on 50 Salads}}
\label{sec:qualitative_50s}
\begin{figure*}[t]
    \centering
    \includegraphics[width=\linewidth]{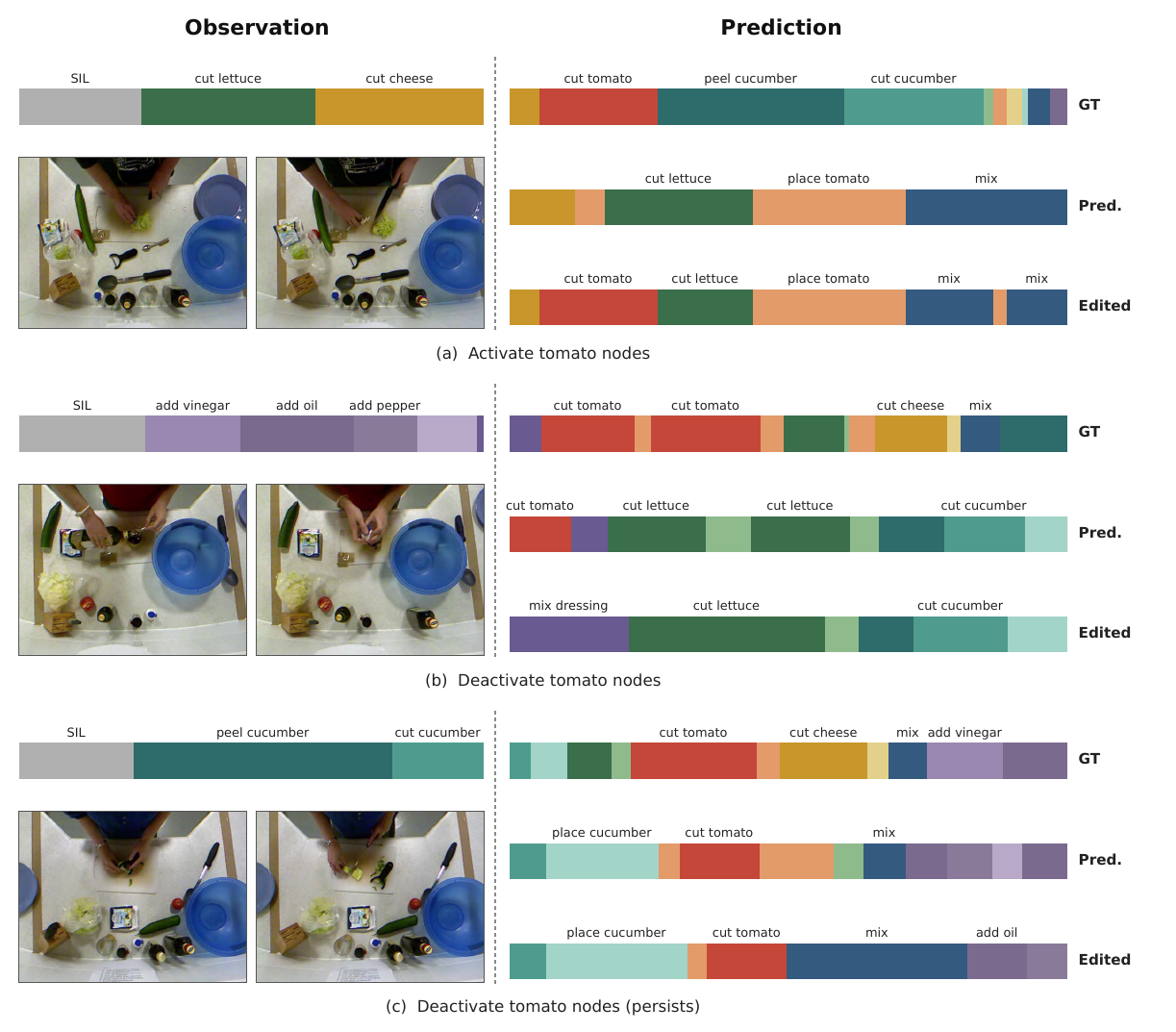}
    \caption{\keeptext{Qualitative anticipation results on 50 Salads with symbolic edits applied at inference. For each example we show the observed segment with two sample frames (left), and the ground-truth future (GT), our prediction (Pred.), and the prediction after editing the tomato nodes of the expanded joint-graph (Edited). (a) the tomato is occluded during the observation and missed by the detector, and activating the nodes recovers the \emph{cut tomato} actions; (b) deactivating them removes cut tomato actions; (c) a case where the prediction persists despite deactivation, i.e.\ the visual stream overrides the symbolic prior.}}
    \label{fig:qual_50s}
\end{figure*}

\keeptext{\Cref{fig:qual_50s} reports qualitative results on 50 Salads, one of the two datasets used for our main comparisons, and simultaneously illustrates the effect of editing the symbolic prior at inference without retraining. Cases (a) and (b) show that activating or deactivating the tomato nodes adds or removes the corresponding \emph{cut tomato} predictions, while case (c) shows a video in which the prediction persists despite deactivation. In case (a), the tomato is partially occluded throughout the observed segment, so the unedited model omits the \emph{cut tomato} actions. A single manual node addition at inference recovers them without retraining. This is the mode of control the symbolic prior is intended to provide, namely correcting a perception failure that a purely neural model offers no handle on.

In case (b), the unedited model initially predicts the \emph{cut tomato} actions. When we manually deactivate the tomato nodes, these predictions are successfully removed. Manual inspection of such cases reveals that the visual evidence for a forthcoming cutting action is often ambiguous. For instance, the tomato is already partially cut during the observed segment. Because the visual stream lacks certainty, the symbolic edit is able to successfully dictate the output.
Case (c) represents the opposite scenario: the action prediction persists despite node deactivation. Here, the tomato is unoccluded and prominent, and the observed actions unambiguously point to cutting as the next step. In this regime, the decoder falls back on the strong visual stream, which accounts for the residual accuracy under deactivation reported in Fig.~3 of the main paper. 
Ultimately, these examples establish a consistent pattern: symbolic edits are highly decisive when the visual observation is incomplete or ambiguous but are sometimes overridden by the partial entanglement of the visual and symbolic pathways (\cref{sec:limitations}) when the visual evidence is strong.}

\section{Limitations and Future work}
\label{sec:limitations}
While our method demonstrates strong performance for action anticipation, it presents certain limitations. First, SymAnt relies on an off-the-shelf object detector to identify task-relevant entities; consequently, initial detection failures can propagate to downstream anticipation. However, because our architecture treats this component as modular, the pipeline will seamlessly benefit from future advancements in visual recognition models. 
Second, the framework depends on a knowledge graph. \droptext{}\keeptext{Although our} KG construction pipeline \droptext{}\keeptext{generally requires little domain engineering, and our ablations show that the model remains robust to general-purpose KGs and new domains,} our future work will focus on \droptext{} automatic knowledge graph extraction directly from foundational models.

Another limitation involves the partial entanglement of visual and symbolic features. As observed in our knowledge editing experiments, manually deactivating an action node drastically reduces, but does not entirely eliminate, the model's ability to predict that action. This indicates a degree of information leakage, where underlying continuous visual features from the video encoder bypass the symbolic bottleneck. While our results prove that modifying the graph still provides significant control over the predicted sequence, future work will focus on strictly disentangling the explicit symbolic logic from the continuous neural feature flow to enable absolute steerability.

\end{document}